\documentclass[letterpaper, 10 pt, conference]{ieeeconf}  %

\IEEEoverridecommandlockouts                              %

\usepackage{booktabs} %
\usepackage{siunitx}  %

\usepackage{cite}
\usepackage{amsmath,amssymb,amsfonts}
\usepackage{algorithmic}
\usepackage{graphicx}
\usepackage{textcomp}
\usepackage{bm}
\usepackage{amsmath}

\usepackage{enumerate}
\usepackage{mathtools}
\usepackage{amssymb}

\usepackage{yhmath}
\usepackage{xcolor}
\usepackage[colorlinks, linkcolor=black, anchorcolor=black, citecolor=black]{hyperref}

\title{\LARGE \bf
Physically-Based Photometric Bundle Adjustment\\ in Non-Lambertian Environments
}
\author{Lei Cheng$^{*}$, Junpeng Hu$^{*}$, Haodong Yan, Mariia Gladkova, Tianyu Huang, \\Yun-Hui Liu, Daniel Cremers, and Haoang Li$^{2,\dagger}$ %
\thanks{Lei Cheng and Junpeng Hu contributed equally to this work. Haoang Li is corresponding author (haoangli@hkust-gz.edu.cn).}
\thanks{This work was supported by the
ERC Advanced Grant SIMULACRON, the Munich Center for Machine
Learning, the EPSRC Programme Grant VisualAI EP/T028572/1, the Shenzhen Portion of Shenzhen-Hong Kong Science and Technology Innovation Cooperation Zone under HZQB-KCZYB-20200089, and the InnoHK of the Government of Hong Kong via the Hong Kong Centre for Logistics Robotics.}
\thanks{L. Cheng, J. Hu, M. Gladkova, and D. Cremers are with Technical University of Munich and Munich Center for Machine Learning, Munich, Germany. H. Yan and H. Li are with The Hong Kong University of Science and Technology (Guangzhou), Guangzhou, China. T. Huang and Y.-H. Liu are with The Chinese University of Hong Kong, Hong Kong, China.}
}

\begin{document}
\maketitle
\thispagestyle{empty}
\pagestyle{empty}

\begin{abstract}
	Photometric bundle adjustment~(PBA) is widely used in estimating the camera pose and 3D geometry by assuming a Lambertian world. However, the assumption of photometric consistency is often violated since the non-diffuse reflection is common in real-world environments. The photometric inconsistency significantly affects the reliability of existing PBA methods.
	To solve this problem, we propose a novel physically-based PBA method. Specifically, we introduce the physically-based weights regarding material, illumination, and light path. These weights distinguish the pixel pairs with different levels of photometric inconsistency.
	We also design corresponding models for material estimation based on sequential images and illumination estimation based on point clouds. In addition, we establish the first SLAM-related dataset of non-Lambertian scenes with complete ground truth of illumination and material.
	Extensive experiments demonstrated that our PBA method outperforms existing approaches in accuracy.
\end{abstract}

\section{INTRODUCTION}

Bundle adjustment is a key component of SLAM~\cite{li2023hong,wang2022efficient}, 3D reconstruction~\cite{huang2023learning}, and structure from motion~\cite{li2020robust}.
By directly aligning the pixels, photometric bundle adjustment (PBA) has demonstrated notable performance in estimating camera poses and reconstructing scene geometry, even in low-texture
environments~\cite{engel2017direct}.

\begin{figure}
	\centering
	\includegraphics[height=0.6\linewidth]{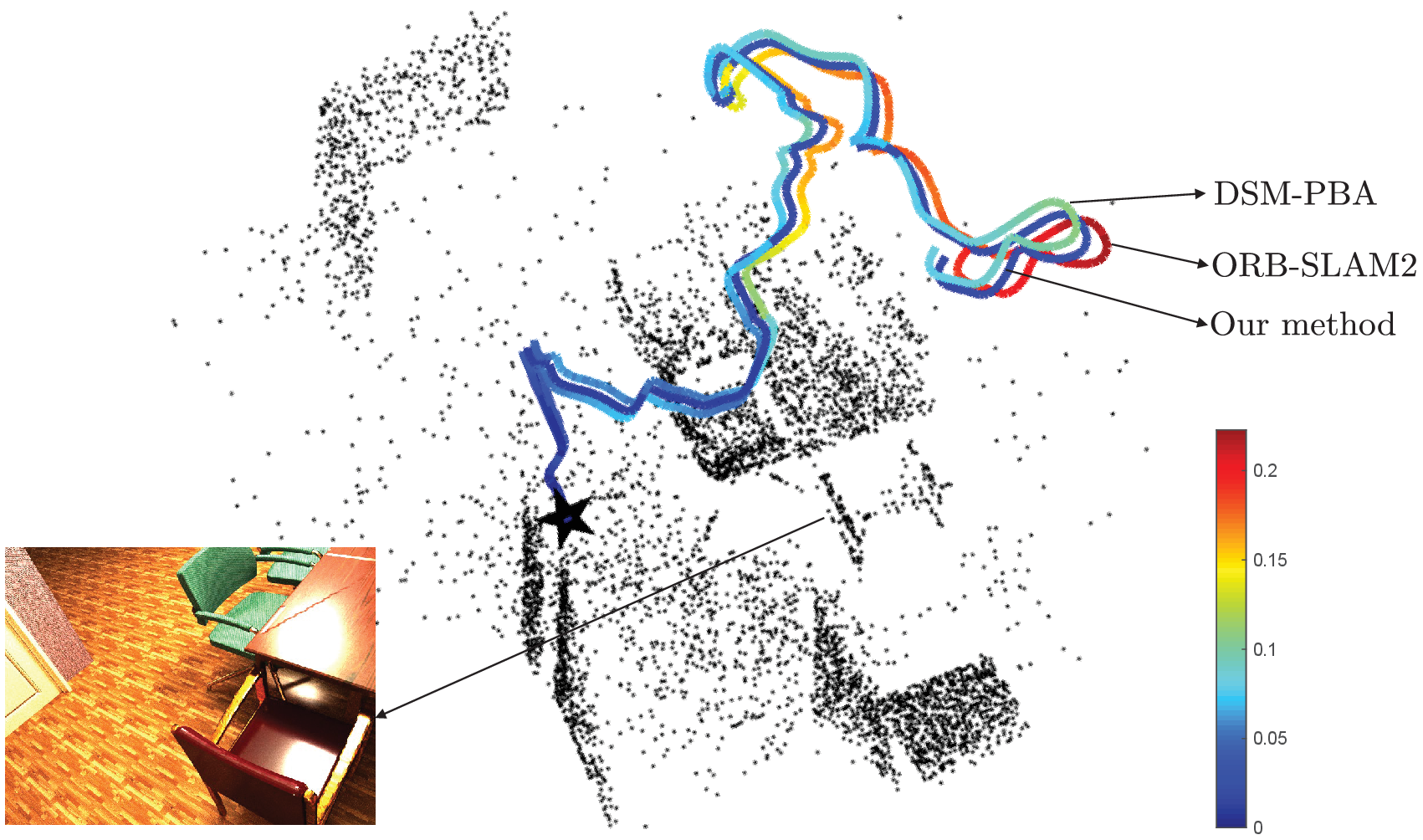}
	\vspace{-2.0em}
	\caption{Effectiveness of our PBA method in a non-Lambertian environment.
		DSM-PBA~\cite{dsm} improves the accuracy of ORB-SLAM2~\cite{mur2017orb}, and our PBA method is more accurate than DSM-PBA. The color bar shows the magnitude of the absolute trajectory error.}
	\vspace{-0.9em}
	\label{fig:1}
\end{figure}

The classic photometric error of PBA is based on the assumption of photometric consistency, implying that corresponding pixels in different views receive identical radiance. However, this assumption holds only for Lambertian surfaces, while non-Lambertian objects are common in real-world situations. These surfaces frequently result in extensive glossy areas, particularly under complex illumination. As illustrated in Fig.~\ref{fig:1}, such glossy regions pose significant challenges to the robustness of existing PBA methods.

Existing strategies of handling photometric inconsistencies can be classified into three categories: invariance-based methods~\cite{zabih1994non, crivellaro2014robust,dai2017bundlefusion,alismail2016direct}, affine model-based methods~\cite{jin2003semi, klose2013efficient,engel2015large, engel2017direct}, and statistical methods~\cite{scandaroli2012improving,meilland2011real,gonccalves2011real,kerl2013robust,dsm}. %
Invariance-based methods, using relative pixel intensity ordering or intensity changes between neighbors, struggle with spatially varying illumination and are sensitive to noise.
Affine model-based approaches often assume a linear relationship between illumination conditions. This simplification is not always applicable in environments with highly reflective surfaces or a wide dynamic range of illumination. Statistical methods like~\cite{kerl2013robust,dsm} can lead to inaccuracies in complex environments where illumination conditions vary locally, inadequately representing outlier distribution caused by photometric inconsistencies. These limitations become pronounced under intensive illumination changes, reducing their effectiveness.

To overcome the above limitations, we propose a novel physically-based photometric bundle adjustment method. We integrate the physically-based reflection model into the classical PBA framework. We design a weighted photometric error to eliminate the impact of pixels corresponding to the non-Lambertian surface points. However, it is challenging to model the weight scalar directly. Based on the physically-based rendering (PBR) model, radiance values are determined by a combination of factors including surface physical properties, scene geometry, and illumination. Therefore, the weight should encode the 1) illumination, 2) object material, and 3) light paths to a point in 3D space. To achieve this, we necessitate material and illumination as prior information. %

Object material can be estimated from images. However, a single image may not be sufficient under challenging illumination conditions. Intuitively, sequential images from multiple viewpoints can resolve potential ambiguities and improve the estimation.
Existing works either focus on single frame input~\cite{barron2013intrinsic, barron2014shape, pbir_2023_ICCV, zhu2022irisformer, li2020inverse}, or necessitate individual optimization on every scene~\cite{jin_tensoir_2023, zhang_modeling_2022, li2022texir}.
Therefore, we develop our new method of material estimation, which processes sequential images with a transformer and outputs the per-pixel roughness, without per-scene optimization. More specifically, we use a transformer to process sequential inputs and capture spatial–temporal correlation over frames.%

Illumination of various positions is required as a prior in our PBA. Moreover, the estimated illumination should be spatially varying and consistent without necessitating per-scene optimization. Existing works either lack a consistent representation~\cite{li2020inverse, zhu2022irisformer} or have a limited field of view~\cite{lighthouse2020}. Therefore, we design our new pipeline of illumination estimation, utilizing a point cloud collected from multiple frames of the scene as input. The output panorama image, termed an ``environment map'', can be used as illumination information for the queried 3D point.

Material and illumination information, serving as priors, will not undergo further optimization in our PBA. However, the light paths are determined by the camera poses and estimated point clouds. Therefore, our weight can be treated as a function of camera poses and point depths. We integrate this term into the classic photometric loss and minimize it to obtain optimal light paths.

Existing SLAM-related datasets~\cite{burri2016euroc, sturm12tumrgbd} do not simultaneously contain sufficient non-Lambertian surfaces and information regarding material and illumination. To overcome their limitations in evaluating non-Lambertian SLAM-related work, we establish our SpecularRooms dataset.

In summary, we make the following contributions:
\begin{itemize}
	\item We propose a novel weighting scheme for photometric error function based on illumination and material information to satisfy the photometric consistency in challenging environments with non-Lambertian surfaces.
	\item We propose new pipelines for material estimation based on sequential images and illumination estimation using point cloud data, without per-scene optimization.
	\item We establish the first PBR dataset of sequential images in non-Lambertian environments with ground truth material and illumination.
\end{itemize}
Our SpecularRooms dataset and supplementary document are available on the project website\footnote{https://sites.google.com/view/haoangli/projects/nlb-pba}.

\section{Related works}

\subsection{Methods for Photometric Consistency}

Depending on the way photometric inconsistencies are handled, we classify existing methods into three primary categories, i.e., illumination invariance-based methods, affine model-based methods, and statistical methods.

\noindent\textbf{Invariance-Based Methods}. Census Transform~\cite{zabih1994non} utilizes relative ordering of intensity, making it invariant concerning monotonic illumination variations.
Alismail et al.~\cite{alismail2016direct} transform images into 8-channel, illumination-invariant binary descriptor images for minimizing photometric error, demonstrating enhanced robustness in subterranean environments. In contrast to the alignment of raw pixels, the approaches presented by Crivellaro et al.\cite{crivellaro2014robust} and Dai et al.~\cite{dai2017bundlefusion}, align gradient magnitudes. This method results in invariance only to local bias changes, making it sensitive to noise and less effective against complex spatially varying illumination.

\noindent\textbf{Affine Model-Based Methods}.  Works by Klose et al.~\cite{klose2013efficient} and Engel et al.~\cite{engel2015large} focus on optimizing the relative pose along with global linear affine brightness transfer parameters between images. DSO\cite{engel2017direct} further incorporates photometric calibration to address the camera's response, vignetting, and exposure variations. Additionally, Jin et al.\cite{jin2003semi} employ affine models for local patches to handle subtle illumination variations, yet this approach falls short under drastic spatial illumination changes.

\noindent\textbf{Statistical Methods}. Some statistics in pixel space are employed to handle photometric inconsistency. Scandaroli et al.~\cite{scandaroli2012improving} employ zero-mean normalized cross correlation in planar template tracking. Median bias normalization, as explored by Meilland et al.~\cite{meilland2011real} and Gonçalves et al.~\cite{ gonccalves2011real}, compensates for intensity biases on both global and patch levels by utilizing the median of residuals. This approach thus achieves partial invariance to variations in intensity bias. To enhance the robustness of PBA,~\cite{dsm} applies a t-distribution to down-weight outliers and achieve state-of-the-art, aiming to mitigate photometric errors. However, this approach falls short of accurately modeling photometric errors under complex illumination conditions, suggesting a need for more sophisticated error modeling techniques.

In contrast, our PBA method considers the physically-based reflection model of each 3D point. By leveraging illumination and material prior information, we can calculate an adaptive weight for every photometric error specifically. Therefore, our method achieves superior performance in chanllenging scenes than previous works.

\subsection{Material and Illumination Estimation}

As mentioned above, our method relies on the estimation of material and illumination. In this part, we shall review related methods.

\noindent\textbf{Material Estimation}. Traditional methods for material estimation, as the works of Barron et al.~\cite{barron2013intrinsic, barron2014shape}, predominantly rely on some strong heuristic priors within an optimization framework. However, these heuristic priors may not be reliable, particularly in indoor scenes characterized by complex geometry and challenging illumination conditions.
Representation-based methods~\cite{jin_tensoir_2023, zhang_modeling_2022, li2022texir} usually take multi-view input and conduct per-scene optimization for their representation. These approaches usually focus on the simultaneous reconstruction of material properties, illumination, and geometry, but rely on per-scene optimization.
Other deep learning-based methods~\cite{pbir_2023_ICCV, li2020inverse}, without per-scene optimization, predominantly focus on single-frame or stereo inputs. %
However, information from a single viewpoint may lead to ambiguity caused by complicated illumination and material.
In contrast, our method, without per-scene optimization, uses sequential images as input to overcome their limitations.

\noindent\textbf{Illumination Estimation}. Early works~\cite{gardner2017learning, sengupta2019neural, legendre2019deeplight} estimate a single environment map from a single image. However, a single environment map is not enough for complex indoor scenes.
Other works~\cite{li2020inverse, zhan2021sparse} consider the spatial variance of illumination, but with a simplified representation. This approximation cannot fully preserve fine-grained illumination information in the observed part of the scene.
Srinivasan et al.~\cite{lighthouse2020} propose a spatial consistent estimation based on multi-plane images and multi-scale volumes. Nonetheless, it takes only stereo inputs, with a limited field of view. Another work~\cite{li2023spatiotemporally} handles spatial and temporal coherence of video input. However, their RNN-based structure limits the capability of representing illumination in long image sequences. %
In our method, we feed point cloud into our model to utilize multi-view information in a consistent way.
\section{Background}

\begin{figure}
	\centering
	\includegraphics[height=0.4
		\linewidth]{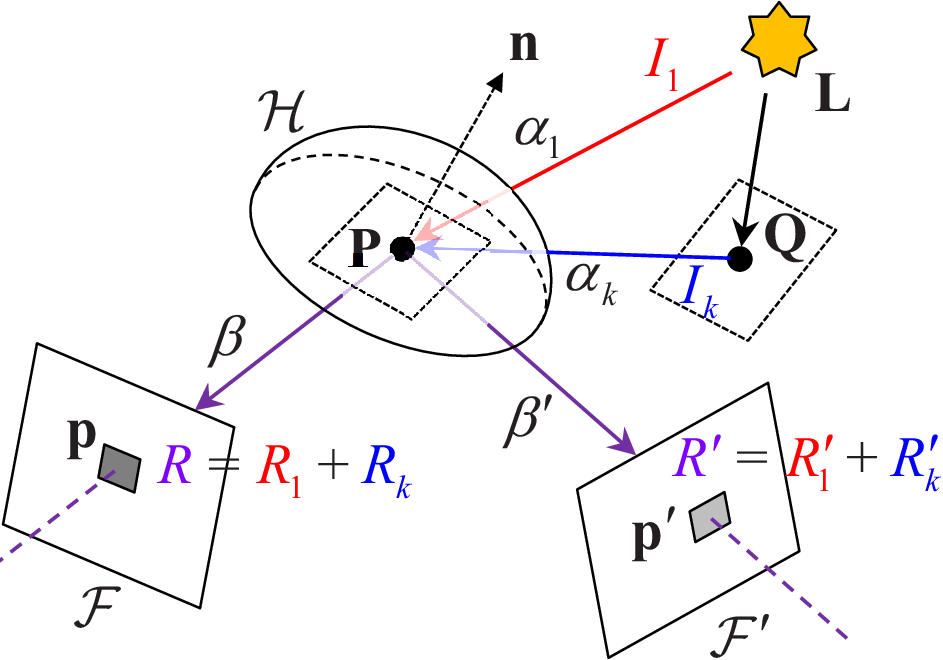}
	\vspace{-1.0em}
	\caption{ Phsically-based reflection model based on point light source. The incident direction $\boldsymbol{\alpha}_k$ is associated with the radiance $I_k$. The radiance received by a 3D point $\mathbf{P}$ is then reflected as the radiance $R_k$ along the reflective direction $\boldsymbol{\beta}$. The pixel $\mathbf{p}$ receives the sum of reflective radiances along the direction $\boldsymbol{\beta}$.}
	\label{fig:radiance}
	\vspace{-0.9em}
\end{figure}

\subsection{Photometric Bundle Adjustment}

\noindent The classical photometric error is formulated as a pixel-wise alignment of two images.
As shown in Fig.~\ref{fig:radiance}, we back-project a pixel~$\mathbf{p} \in \mathbb{R}^{2}$ from the reference frame $\mathcal{F}$ into a 3D point~$\mathbf{P} \in \mathbb{R}^{3}$ using the depth~$d$ of $\mathbf{p}$. Then we project the 3D point~$\mathbf{P}$ into a pixel~$\mathbf{p}^{\prime}$ on the $i_{th}$ visible frame $\mathcal{F}^{\prime}$ using the unknown rotation~$\mathbf{R}$ and translation~$\mathbf{t}$ between two frames. Accordingly, the pixel~$\mathbf{p}^{\prime}$
is with respect to the pose~$\mathbf{R},\mathbf{t}$ and the depth~$d$. With using the pinhole model and given camera intrinsics  in $\mathbf{p}^{\prime}(\mathbf{R}, \mathbf{t}, d)$ implicitly, the classic photometric error is defined by

\begin{equation}
	E_{\mathbf{p} i} = \sum_{ \mathbf{p} \in \mathcal{N}_{\mathbf{p}} } \left\|\mathcal{F}(\mathbf{p})-
	\mathcal{F}^{\prime}\left(\mathbf{p}^{\prime}(\mathbf{R}, \mathbf{t}, d)\right)    )\right\|,
	\label{eq:ssd_pixel}
\end{equation} where~$\mathcal{N}_\mathbf{p}$ denote a set of pixels around the pixel~$\mathbf{p}$.
This photometric error is predicted on the assumption that the 3D point~$\mathbf{P}$ resides on a Lambertian surface, where the radiance, power received by pixels~$\mathbf{p}$ and $\mathbf{p}^{\prime}$, is uniform across viewpoints. However, for non-Lambertian objects, their surface reflection varies with viewing direction intensively.

Fundamentally, PBA optimizes scene geometry and camera poses jointly by minimizing the photometric error of projecting all estimated 3D points on their visible frames. The classical PBA loss is given by

\begin{equation}
	\mathcal{L}_\text{photo} =\sum_{j \in \mathcal{F}} \sum_{\mathbf{p} \in \mathbf{P}_j} \sum_{i \in \operatorname{obs}(\mathbf{p})} E_{\mathbf{p} i} \text {,}
	\label{eq:pba_loss_dso}
\end{equation} where $j$ traverses through all frames $\mathcal{F}$, and within each frame $j$, $\mathbf{p}$ iterates over all points $\mathbf{P}_j$. The $\operatorname{obs}(\mathbf{p})$ denotes all visible frames for $\mathbf{p}$.

\subsection{Physically-Based Reflection Model}

\label{sec:secb}
To formulate the PBA with the physically-based reflection model, we start with a conceptual exposition using direct lighting. As illustrated in Fig.~\ref{fig:radiance}, at a 3D point~$\mathbf{P}$, an incident light direction~$\bm{\alpha}_1$ is associated with the radiance~$I_1$. The ``valid'' radiance received by $\mathbf{P}$ is the component along the surface normal~$\mathbf{n}$, i.e., $I_1 \cdot \cos \theta_{\bm{\alpha}_1,\mathbf{n}} $. For the surface around the point~$\mathbf{P}$, we denote its normal by~$\mathbf{n}$. After reflection at point~$\mathbf{P}$, the valid radiance is distributed to different reflective directions such as~$\bm{\beta}$ and~$\bm{\beta}^{\prime}$.
Bidirectional reflectance distribution function (BRDF)~\cite{Cook1982}~$f_{\mathbf{p}}$ determines the distributed radiances~$r_1$ and $r_1^{\prime}$ along the directions~$\bm{\beta}$ and~$\bm{\beta}^{\prime}$, i.e.,
\begin{subequations}
	\label{eq:two_radiances_simp}
	\begin{align}
		r_1 = I_1 \cdot \cos\theta_{\bm{\alpha}_1, \mathbf{n}} \cdot  f_{\mathbf{P}}( \bm{\alpha}_1, \mathbf{n}, \bm{\beta} ) 	,
		\label{subeq:R1}
		\\
		r_1^{\prime} = I_1 \cdot \cos\theta_{\bm{\alpha}_1, \mathbf{n}} \cdot  f_{\mathbf{P}}( \bm{\alpha}_1, \mathbf{n}, \bm{\beta}^{\prime} ).
	\end{align}
\end{subequations}

\section{Our Method}

Given an initial estimate of the scene points~$\left\{\mathbf{P}_j\right\}_{j=1}^N$ and viewing parameters $\left\{\mathbf{T}_i\right\}_{i=1}^M$ per frame in the world, we first estimate each pixel's roughness and normal in every frame. Meanwhile, we get the colored point cloud of the scene and select control points. Then the illumination network predicts environment maps of control points, with the point cloud as input.
Leveraging environment maps, surface roughness, and surface normals, we optimize the trajectory and the scene geometry in our PBA method.

\subsection{PBA with a Physically-Based Reflection Model}

In this section, we introduce a weighted photometric error in PBA loss, which is defined as

\begin{equation}
	\mathcal{L} =\sum_{j \in \mathcal{F}} \sum_{\mathbf{p} \in \mathbf{P}_j} \sum_{i \in \operatorname{obs}(\mathbf{p})} \delta_{\mathbf{p}i} \cdot E_{\mathbf{p} i} \text {,}
	\label{eq:pba_loss_new}
\end{equation}

\noindent where we define the weight as
\begin{equation}
	\delta_{\mathbf{p}i} = \mathrm{e}^{-\theta |r - r^{\prime}|}.
	\label{eq:inv_weight_func}
\end{equation}
\noindent Here, $r$ and $r'$ correspond to radiance reflected by 3D point $\mathbf{p}$ in two view directions.
In static scenes, unoccluded points with significant variations in reflected radiance result in large photometric errors. To decrease the impact of these points, we propose a straightforward inversely proportional function to determine the weight $\delta$ in Eq.~(\ref{eq:inv_weight_func}). We empirically found that $\theta = 14.6$ leads to superior PBA performance. Ablation studies are available in the supplementary document.

In practice, the calculation of radiance $r$ in Eq.~(\ref{eq:inv_weight_func}) should consider both direct and indirect lighting.
To define our radiance model, we first introduce the global radiance received
by the pixel~$\mathbf{p}$. As shown in Fig.~\ref{fig:radiance}, each incident direction~$\bm{\alpha}_k$ is associated with the radiance~$I_k$. The unit hemisphere~$\mathcal{H}$ covers all the incident directions.  Analogous with Eq.~(\ref{subeq:R1}), the reflective radiance along with the reflective direction~$\bm{\beta}$ becomes $I_k \cdot \cos \theta_{\mathbf{n}, \bm{\alpha}_k}	 \cdot f_{\mathbf{P}}( \bm{\alpha}_k, \mathbf{n}, \bm{\beta})$.
By integrating the radiance functiontion over the hemisphere~$\mathcal{H}$, we compute
the total radiance $r$ received by the pixel~$\mathbf{p}$ as
\begin{equation}
	r = \int_{\mathcal{H}} I_k \cdot \cos \theta_{\mathbf{n}, \bm{\alpha}_k}	 \cdot f_{\mathbf{P}}( \bm{\alpha}_k, \mathbf{n}, \bm{\beta}) \ \mathrm{d} \bm{\alpha}_k.
	\label{eq:orig_int}
\end{equation}

To represent the physical reflection process, i.e., the light path, we detail the formulation of the vectors~$\mathbf{n}$, $\bm{\beta}$, and $\bm{\beta}^{\prime}$ from Eq.~(\ref{eq:orig_int}) in the following subsections.

\noindent\textbf{Normal of Facet~$\mathbf{n}$.} We back-project the pixel~$\mathbf{p}$ and its neighbor pixels using the measured depths~$d$ and its neighbouring depth~$\{ d_i \}$
from the depth map, obtaining 3D points~$\mathbf{P}(d)$ and its surrounding points within a 2cm vicinity~$\{\mathbf{N}_i(d_i)\}$.
Each 3D point triplet, e.g.,~$\{\mathbf{P}(d)$, $\mathbf{N}_1(d_1)$, $\mathbf{N}_2(d_2)\}$ leads to a normal, e.g.,
$\mathbf{n}(d,d_1,d_2) = \big( \mathbf{P}(d)-\mathbf{N}_1(d_1) \big) \times \big( \mathbf{P}(d)-\mathbf{N}_2(d_2) \big) $. Then we use the moving least square method\cite{levin1998approximation} to smooth the estimated normal.

\noindent\textbf{Reflective Direction~$\bm{\beta}$ and $\bm{\beta}^{\prime}$.} With the relative rotation~$\mathbf{R}$ and translation~$\mathbf{t}$, the reflective directions~$\bm{\beta}$ and $\bm{\beta}^{\prime}$ can be expressed as
\begin{align}
	\bm{\beta}(d)                                & = \mathbf{P}(d)\text{,}                             \\
	\bm{\beta}^{\prime}(\mathbf{R},\mathbf{t},d) & = \mathbf{P}(d)+\mathbf{R}^{\top}\mathbf{t}\text{.}
\end{align}

\noindent Accordingly, we re-formulate the photometric loss~$\mathcal{L}_\text{photo}$ (Eq.~\eqref{eq:pba_loss_dso}) as
\begin{equation}
	\mathcal{L} =\sum_{j \in \mathcal{F}} \sum_{\mathbf{p} \in \mathbf{P}_j} \sum_{i \in \operatorname{obs}(\mathbf{p})} \delta_{\mathbf{p}i}(\mathbf{R}, \mathbf{t}, d,\left\{d_i\right\}) \cdot E_{\mathbf{p} i} \text {,}
	\label{eq:new_loss_direct}
\end{equation}
which additionally incorporates a physically-based reflection model.
Nevertheless, the proposed loss~$\mathcal{L}$ cannot be directly minimized using a gradient descent algorithm since the partial derivatives of several integrals in the loss over the camera parameters~$\big\{ \mathbf{R}, \mathbf{t}, d,\{d_i\} \big\}$ are difficult to compute~\cite{li2018monocular}. To avoid computing these integrals in Eq.~(\ref{eq:orig_int}), we  leverage the estimated environment lighting and roughness in BRDF~\cite{Cook1982}.

Given the surface material, reflective direction~$\bm{\beta}$ and surface normal~$\mathbf{n}$, we can determine
the solid angle area~$\mathcal{A}_{\mathbf{P}}(\mathbf{n}, \bm{\beta})$ in BRDF. %
Hence, we split
the irradiance~$r$ in Eq.~(\ref{eq:orig_int}) as
\begin{equation}
	r \approx
	\underbrace{
	\dfrac{ 	\int_{\mathcal{A}_{\mathbf{P}}(\mathbf{n},\bm{\beta})}
	I_k  \	\mathrm{d} \bm{\alpha}_k }
	{  \int_{\mathcal{A}_{\mathbf{P}}(\mathbf{n},\bm{\beta})} 	\mathrm{d} \bm{\alpha}_k	}
	}_{T_{\textnormal{Light}}}
	\cdot
	\underbrace{ \int_{\mathcal{H}}
		\cos \theta_{\mathbf{n}, \bm{\alpha}_k} \cdot
		f_{\mathbf{P}}( \bm{\alpha}_k, \mathbf{n}, \bm{\beta})  \	\mathrm{d} \bm{\alpha}_k }_{T_{\textnormal{BRDF}}}.
	\label{eq:split_integral}
\end{equation}

\begin{itemize}
	\item \textbf{Term~$\mathbf{T_\text{Light}}$.}
	      Given the initial structure and estimated cameras,
	      we use a viewing direction $\bm{\beta}$ and $\mathbf{n}$ to
	      compute the area~$\mathcal{A}_{\mathbf{P}}$ and to further index the pre-computed integral of the environment map.

	\item \textbf{Term $\mathbf{T_\text{BRDF}}$.} Term $T_{\textnormal{BRDF}}$ is predominantly influenced by the surface roughness $r_s$ of the surface material, which ranges from 0 to 1.
\end{itemize}
\noindent\textbf{Illumination at Control Points}.
According to Eq.~(\ref{eq:orig_int}), we need environment lighting $I$ for every point in PBA. However, it is impractical and unnecessary to predict all of them, as points in the vicinity are under similar illumination conditions. Therefore, we select representative ``control points" on the surfaces by employing SLIC~\cite{achanta2012slic} clustering
on the normal maps. Subsequently, we estimate the environment maps for these control points as introduced in Section \ref{illumination Estimation}.
For each 3D point, we employ the predicted environment map of its nearest ``control point" in Eq.~(\ref{eq:orig_int}).

Utilizing the estimated roughness $r_s$, as detailed in Section \ref{Material Estimation}, along with environment lighting for each point, we compute the specular radiance $r$ of 3D points across all views. Subsequently, the radiance is applied in Eq.~(\ref{eq:inv_weight_func}) to weigh each photometric error present in the loss function $\mathcal{L}$. To optimize Eq.~(\ref{eq:new_loss_direct}), initial camera poses and scene geometry can be obtained within SLAM frameworks. During optimization, the refined camera poses and geometry informs the light path determination in subsequent optimizations. We follow~\cite{li2019leveraging} to use the Levenberg-Marquardt algorithm
to minimize the weighted PBA loss.

\subsection{Material Estimation\label{Material Estimation}}

\begin{figure}[t!]
	\centering
	\includegraphics[width=\linewidth]{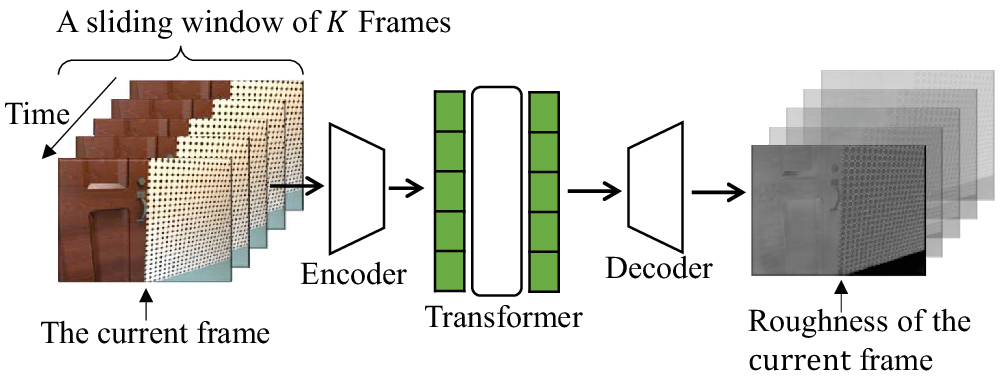}
	\vspace{-2.0em}
	\caption{Material estimation pipeline. Our approach uses the current frame and $K-1$ previous frames to predict the roughness of the current frame. A CNN encoder extracts features from these $K$ frames.  A transformer and a decoder process these features to predict the roughness of the current frame.}
	\label{fig_mat_structure}
	\vspace{-0.9em}
\end{figure}

The appearance of surface material varies significantly under different illumination conditions, leading to ambiguities in material estimation from single-frame inputs.
To handle these ambiguities, we use the multi-view information from sequential images. Our material estimation network leverages previous $K-1$ frames to enhance the estimation of the current frame. %
As shown in  Fig.~\ref{fig_mat_structure}, the architecture of our material estimation model consists of three principal components: (1) a CNN encoder to extract features from individual frames, (2) a transformer to capture and model the spatial–temporal relationships across frames, and (3) a CNN decoder to produce the per-pixel roughness.

\noindent\textbf{Single-Frame Encoder}. Our model adopts the CNN encoder structure as described in \textsf{IRCIS}~\cite{li2020inverse}. Initially, frame-level features are extracted for each frame using this encoder. These extracted features from $K$ frames are then concatenated to construct a multi-frame
feature map, denoted as $f \in \mathbb{R}^{K \times C \times H' \times W'}$.

\noindent\textbf{Spatial–temporal Correlation}. We use a transformer to capture the spatial–temporal correlations among different locations across frames. We flatten the feature maps of $K$ frames into $f \in \mathbb{R}^{d \times (K \cdot H^\prime \cdot W^\prime)}$, which are $K \cdot H^\prime \cdot W^\prime$ tokens of dimension $d$. These tokens are fed into the transformer, with temporal and spatial positional information encoded by 3D position embedding as in~\cite{wang2021end}. %

\noindent\textbf{Output Decoder}. We adopt the CNN decoder structure from \textsf{IRCIS}~\cite{li2020inverse} to output per-pixel roughness. %

\noindent\textbf{Loss Function}. During training, the model estimates roughness maps of $i-K$ to $i$ frames. A mean squared error (MSE) loss term ${L}_\textnormal{MSE}$ measures pixel-level error and a structural similarity index measure (SSIM)~\cite{wang2004ssim} term ${L}_\textnormal{SSIM}$ assesses the similarity of the overall structure. The final loss function ${L_\textnormal{mat}}$ is a weighted sum of the two terms:
\begin{equation}
	{L_\textnormal{mat}} = \lambda_\textnormal{MSE} \cdot {L}_\textnormal{MSE} + \lambda_\textnormal{SSIM} \cdot {L}_\textnormal{SSIM},
\end{equation}
where $\lambda_\textnormal{MSE}$ and $\lambda_\textnormal{SSIM}$ are the corresponding weights of MSE loss and SSIM loss.

\subsection{Illumination Estimation \label{illumination Estimation}}

\begin{figure}[t!]
	\centering
	\includegraphics[width=\linewidth]{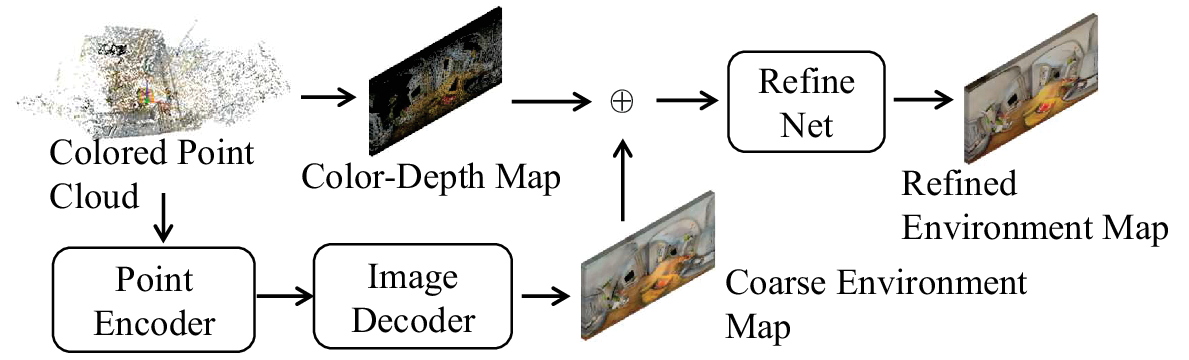}
	\vspace{-2.0em}
	\caption{
		Illumination estimation pipeline. The colored point cloud is fed into the point encoder, which extracts per-point features. These features are then projected and used as inputs for the image decoder. The image decoder then produces a coarse output. The coarse output is combined with a color-depth map in the RefineNet, yielding a refined output.    }
	\label{fig:light_architecture}
	\vspace{-0.9em}
\end{figure}

For every control point in the scene, its illumination varies with its position, especially in indoor scenes with complex occlusion and illumination. Therefore we need to estimate a spatially varying illumination of the whole scene. Compared to single-image input, multi-view images of the scene brings a larger field of view, which benefits this task. However, individually processing these images might lead to spatial inconsistency, since they are from different viewpoints. To avoid this inconsistency, we need to integrate those images into a unified 3D representation. In our PBA framework, a colored point cloud of the scene is available and we utilize it as input for our illumination prediction pipeline.
Our illumination estimation pipeline is shown in Fig.~\ref{fig:light_architecture}. Given the scene's point cloud, the model outputs an environment map as the environment lighting of the queried position.%

\begin{figure*}[htbp]
	\footnotesize
	\centering
	\renewcommand{\tabcolsep}{3.0pt}
	\begin{tabular}{cc}
		$\mathsf{ORB}$-$\mathsf{SLAM}$2~\cite{mur2017orb}
		 &
		\begin{tabular}{ccc}
			\emph{``Office"}                                                                                & \emph{``Computer Room"} & \emph{``Meeting Room"}
			\\
			\hline                                                                                                                                             \\[-0.5em]
			\includegraphics[height=0.21\linewidth]{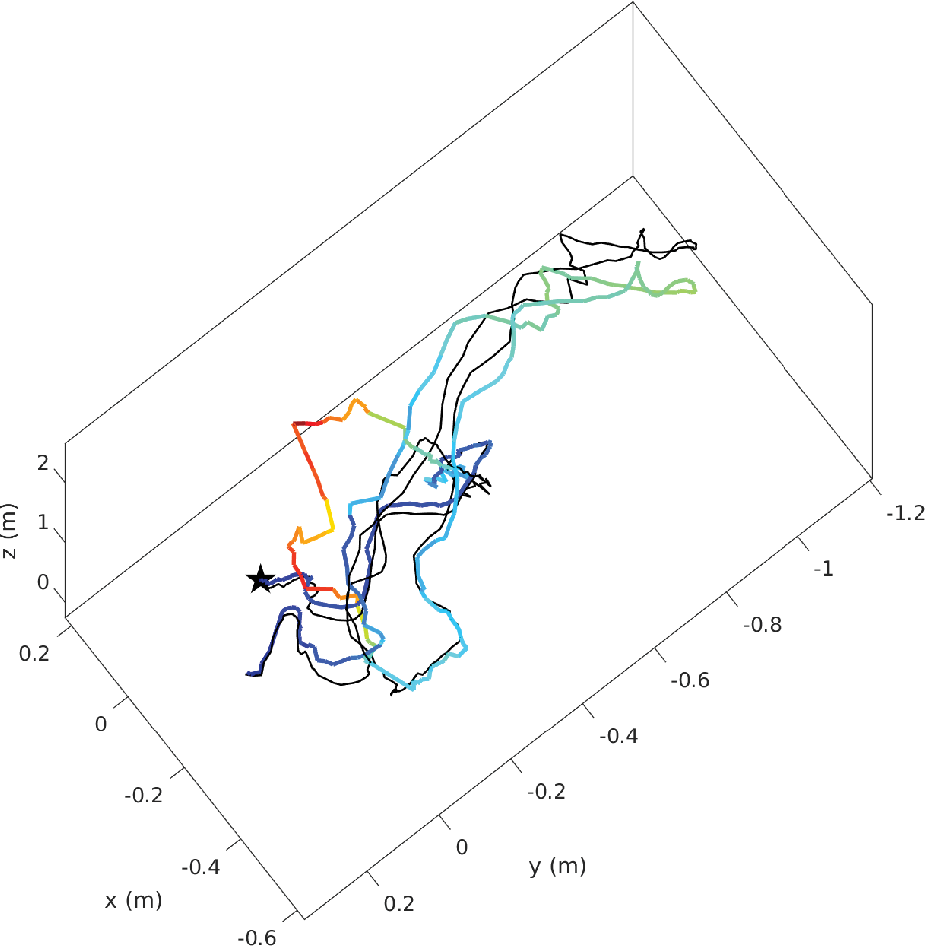} &
			\includegraphics[height=0.21\linewidth]{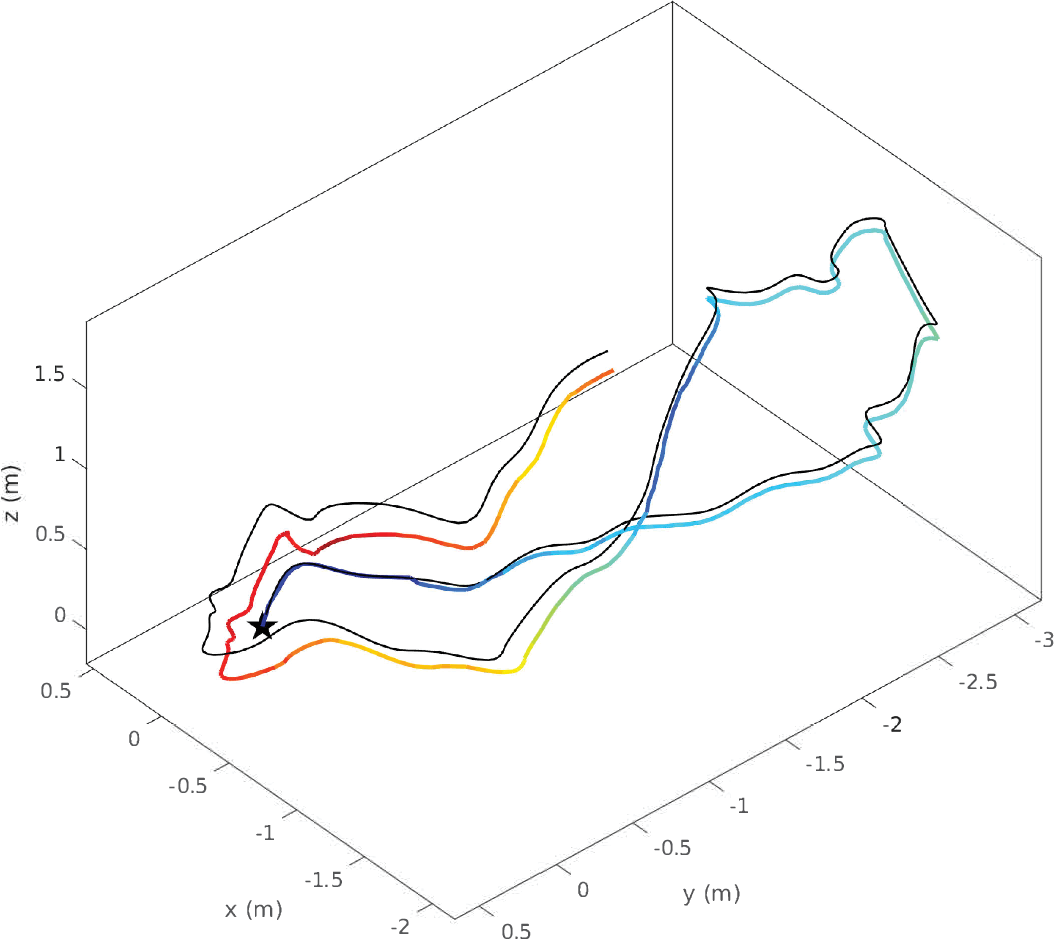}                   &
			\includegraphics[height=0.21\linewidth]{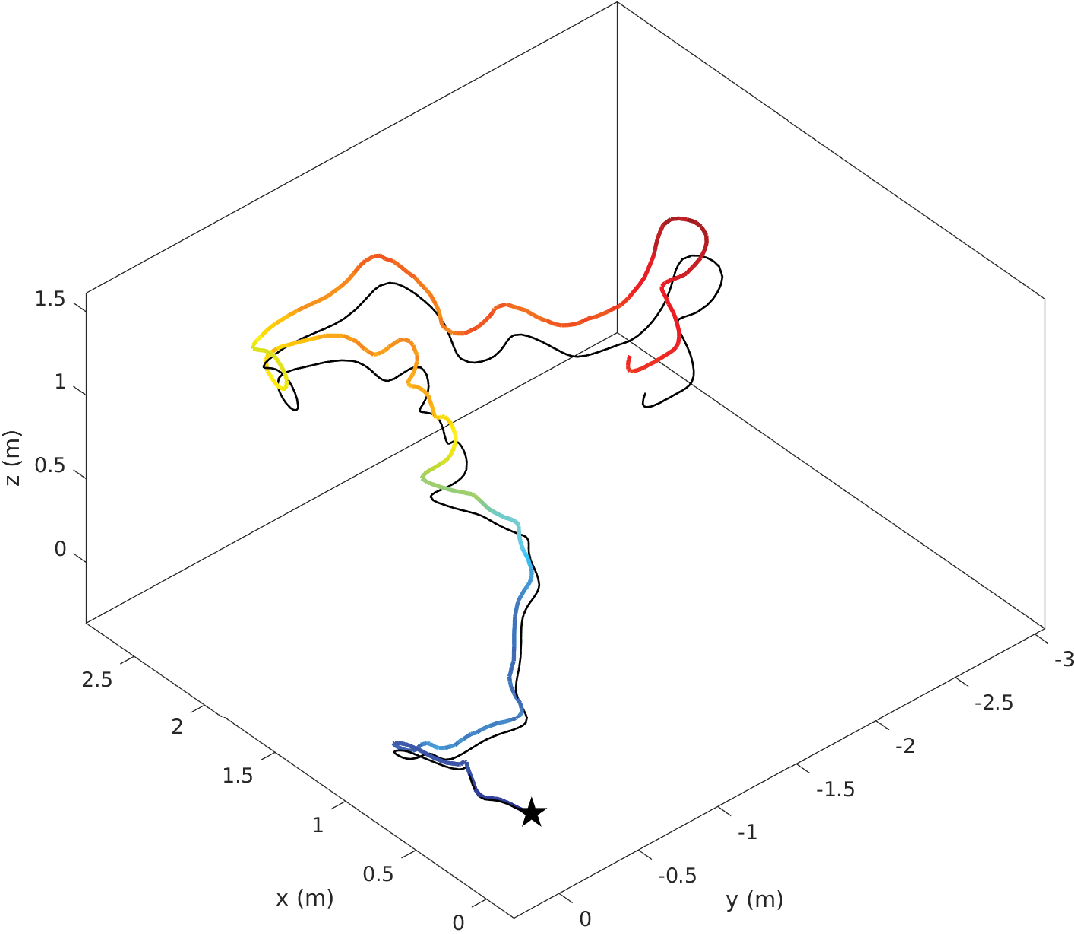}
			\\[0.3em]
		\end{tabular}
		\\
		$\mathsf{LBT}$-$\mathsf{PBA}$~\cite{pba_cmu}
		 &
		\begin{tabular}{ccc}
			\includegraphics[height=0.21\linewidth]{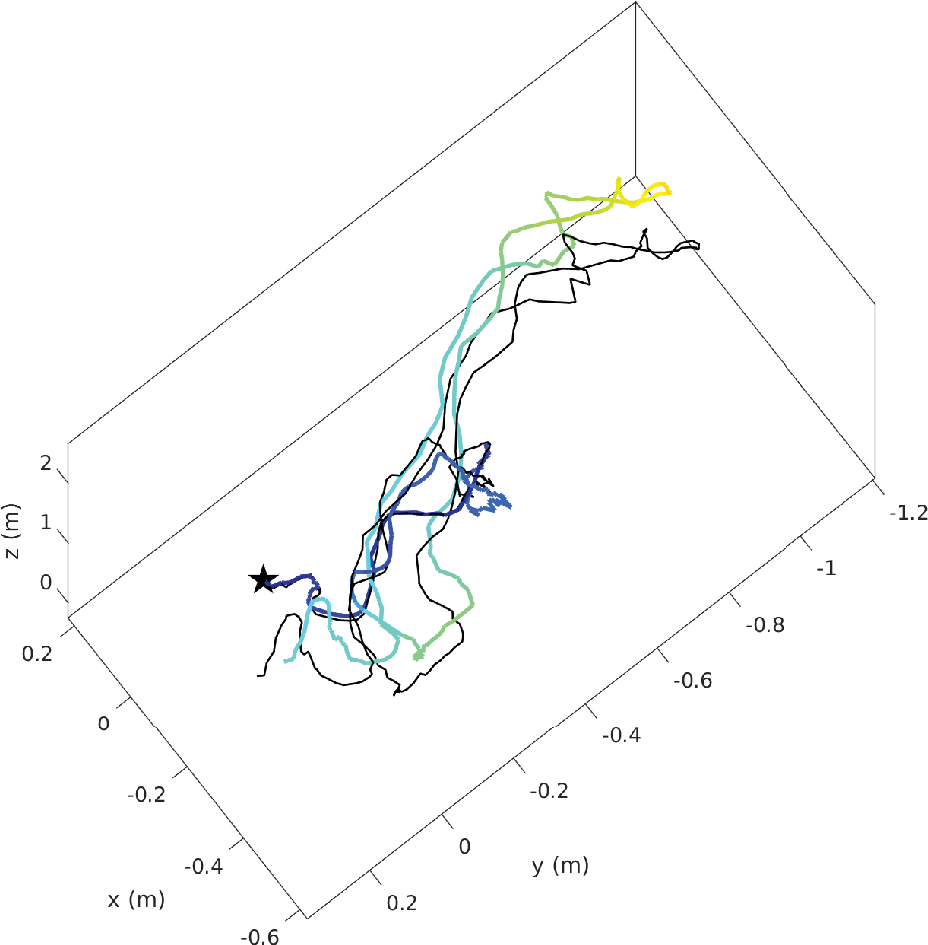} &
			\includegraphics[height=0.21\linewidth]{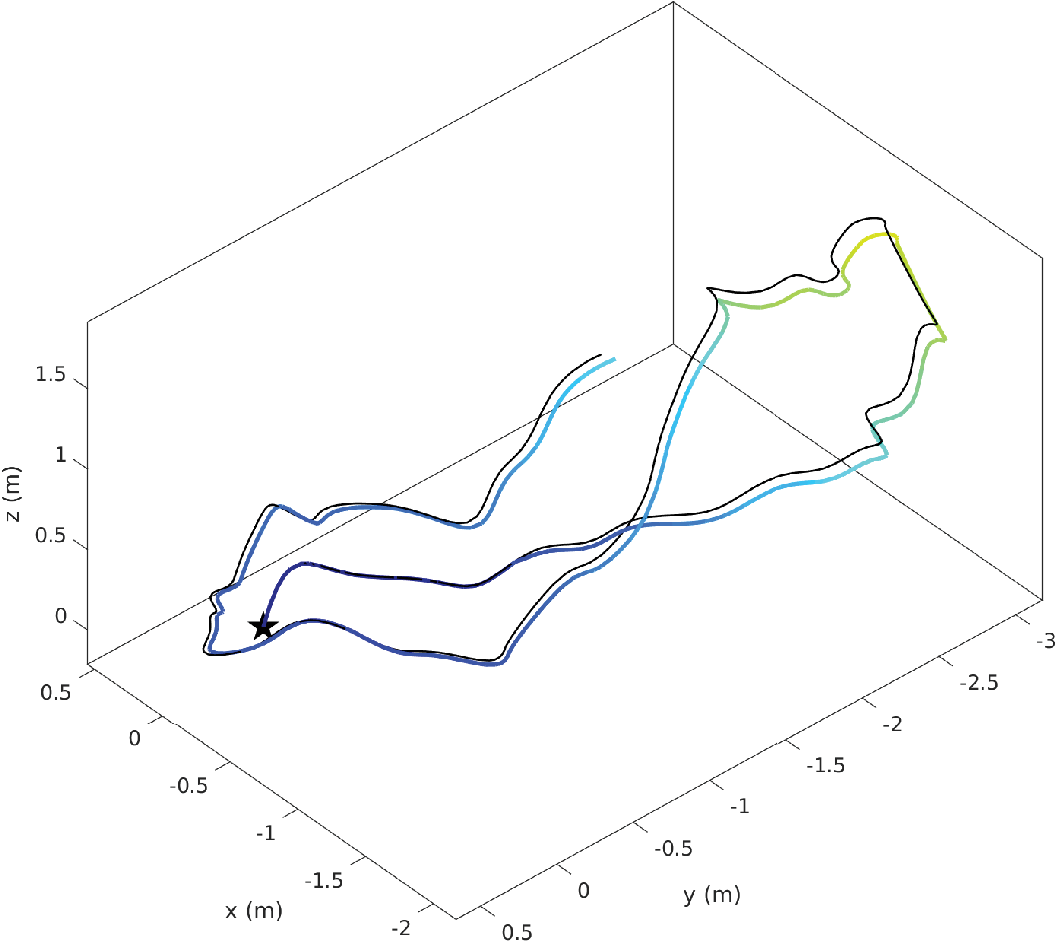}                   &
			\includegraphics[height=0.21\linewidth]{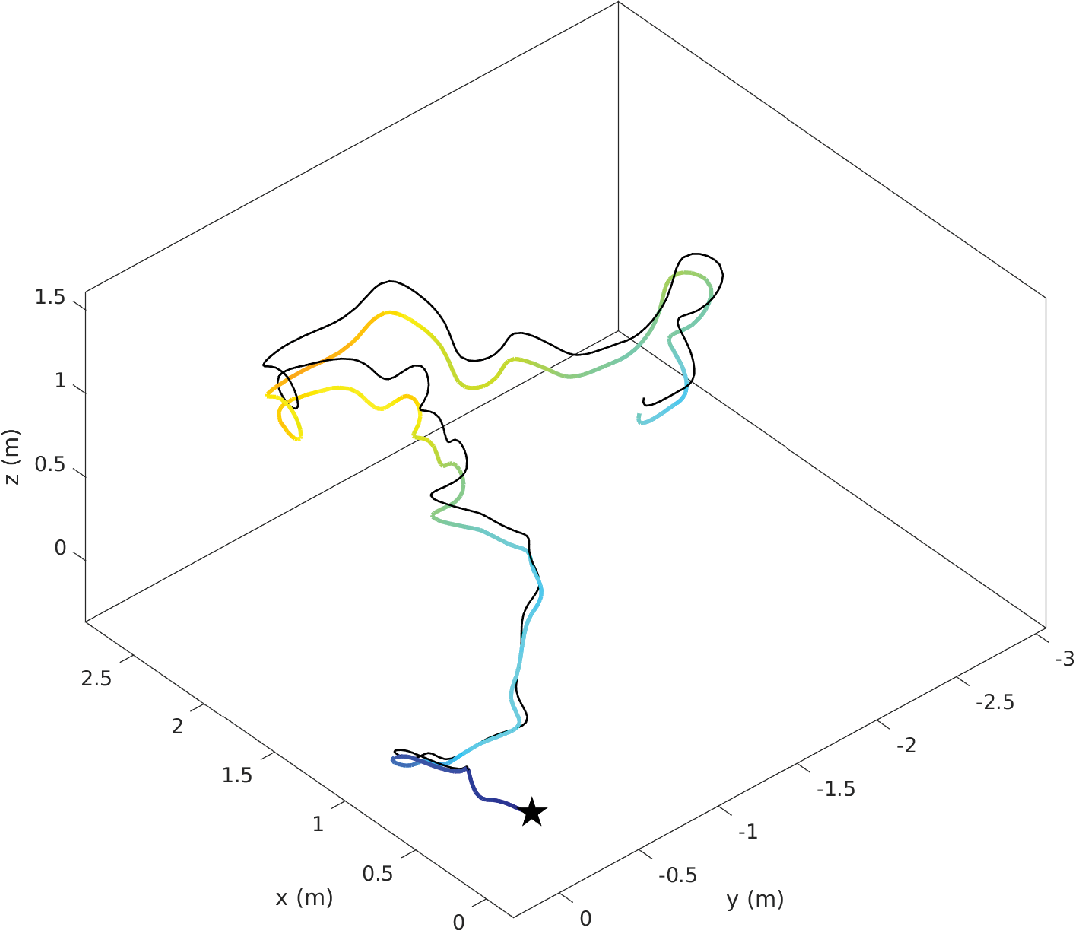}
			\\[0.3em]
		\end{tabular}
		\\
		$\mathsf{DSM}$-$\mathsf{PBA}$~\cite{dsm}
		 &
		\begin{tabular}{ccc}
			\includegraphics[height=0.21\linewidth]{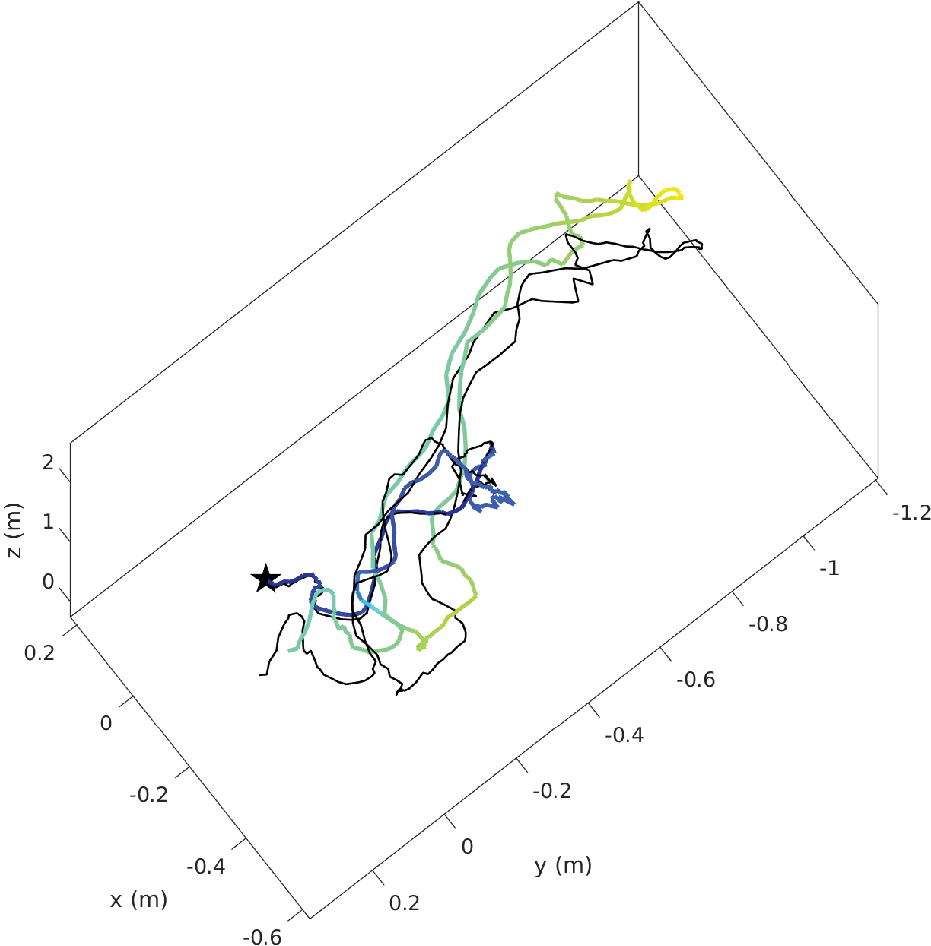} &
			\includegraphics[height=0.21\linewidth]{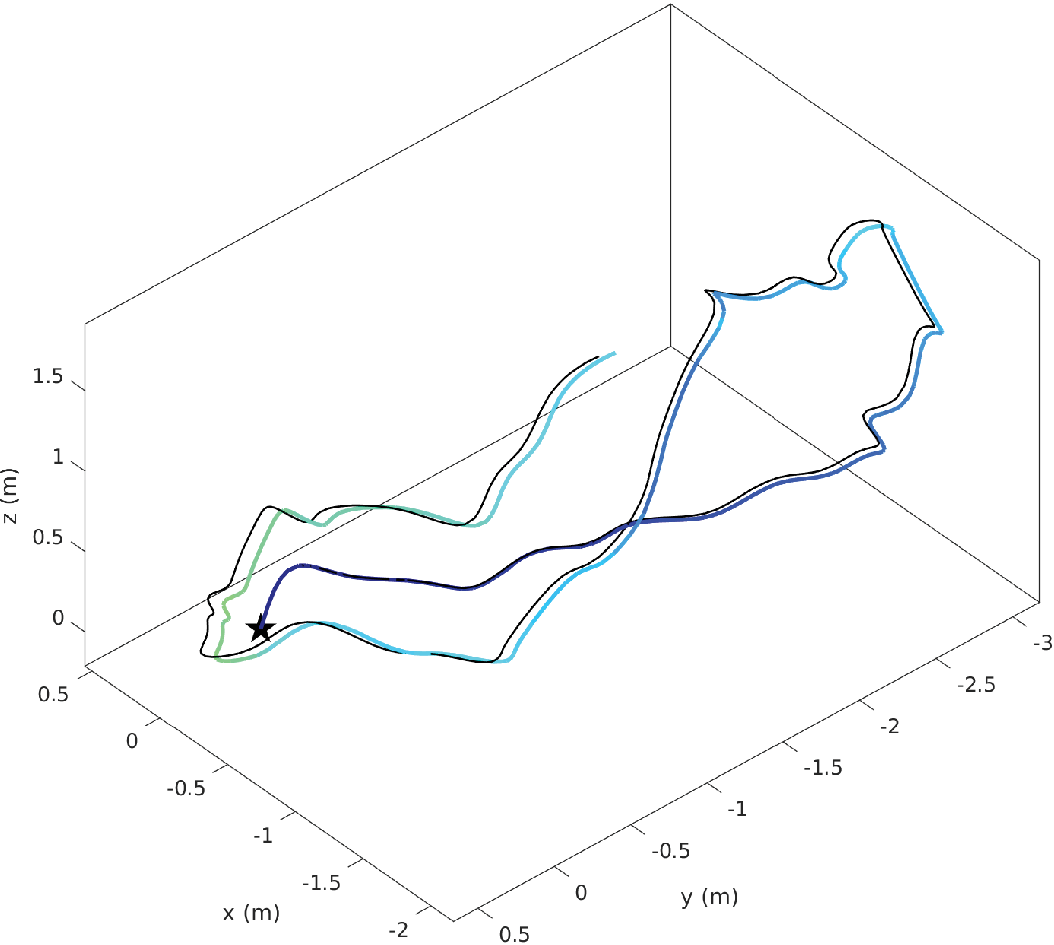}                   &
			\includegraphics[height=0.21\linewidth]{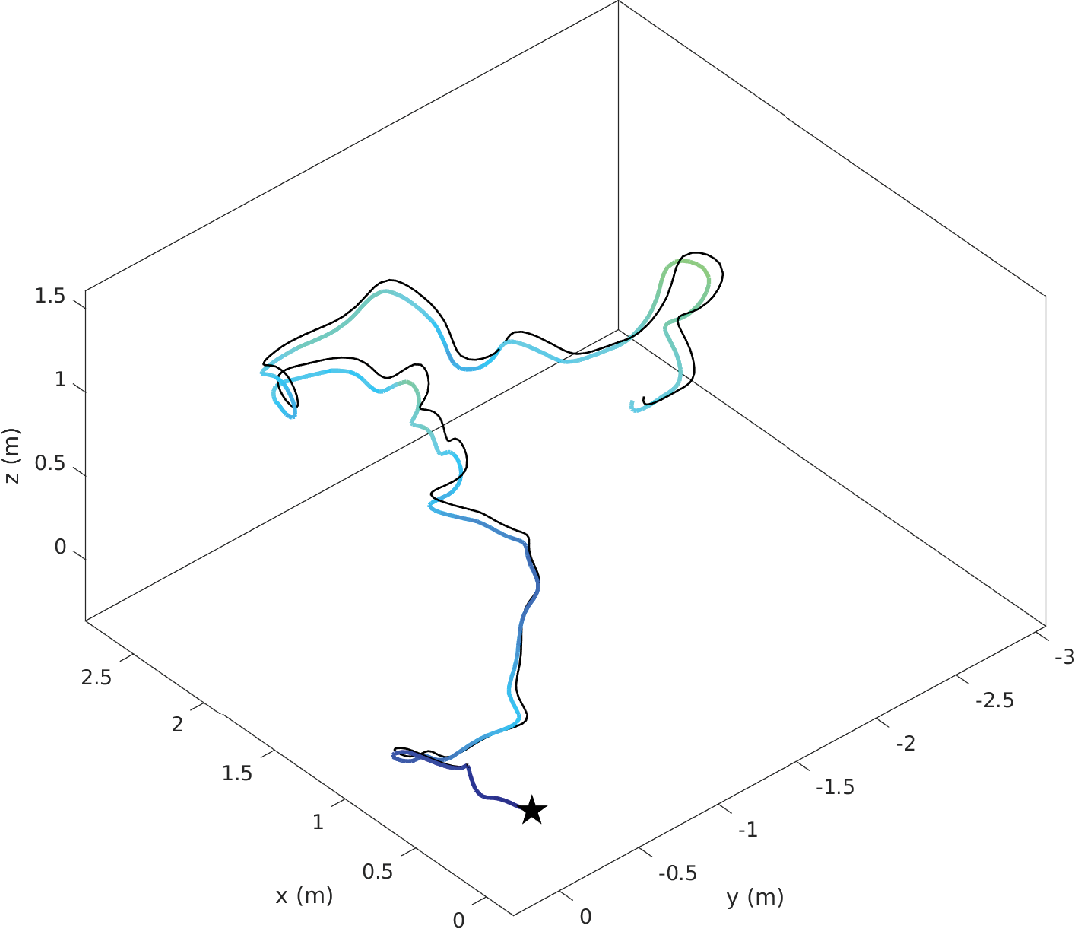}
			\\[0.3em]
		\end{tabular}
		\\
		$\mathsf{PB}$-$\mathsf{PBA}$~(our)
		 &
		\begin{tabular}{ccc}
			\includegraphics[height=0.21\linewidth]{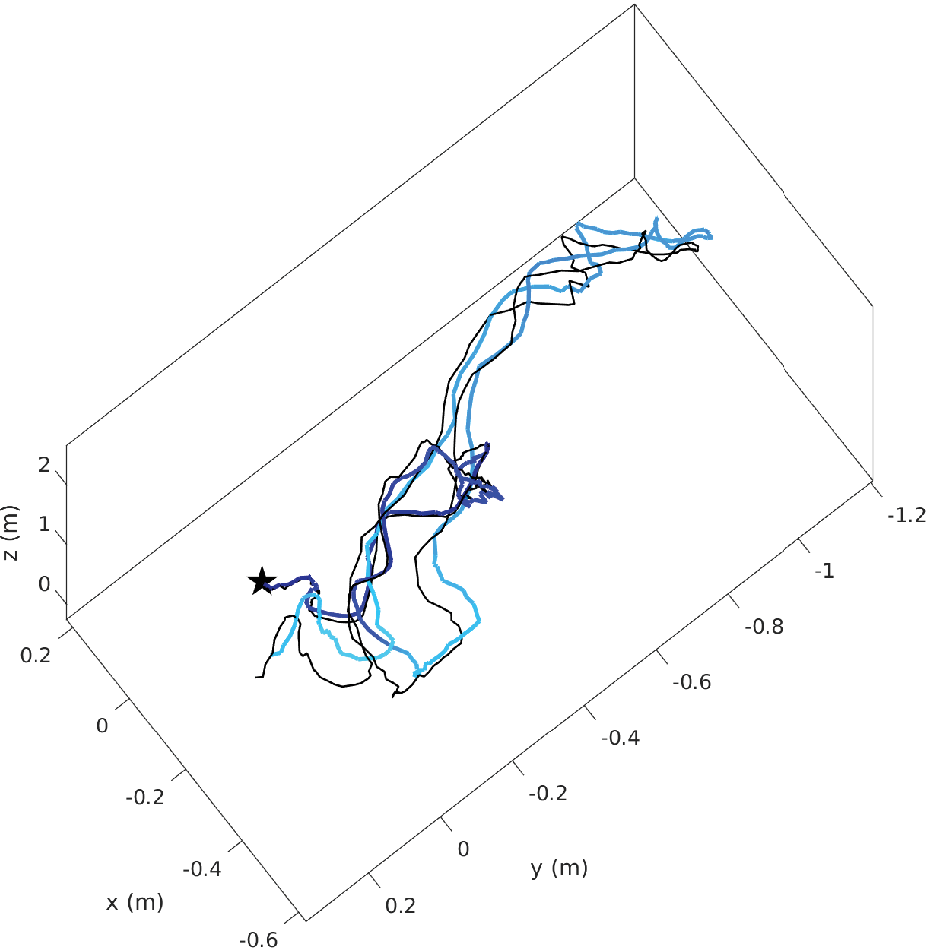} &
			\includegraphics[height=0.21\linewidth]{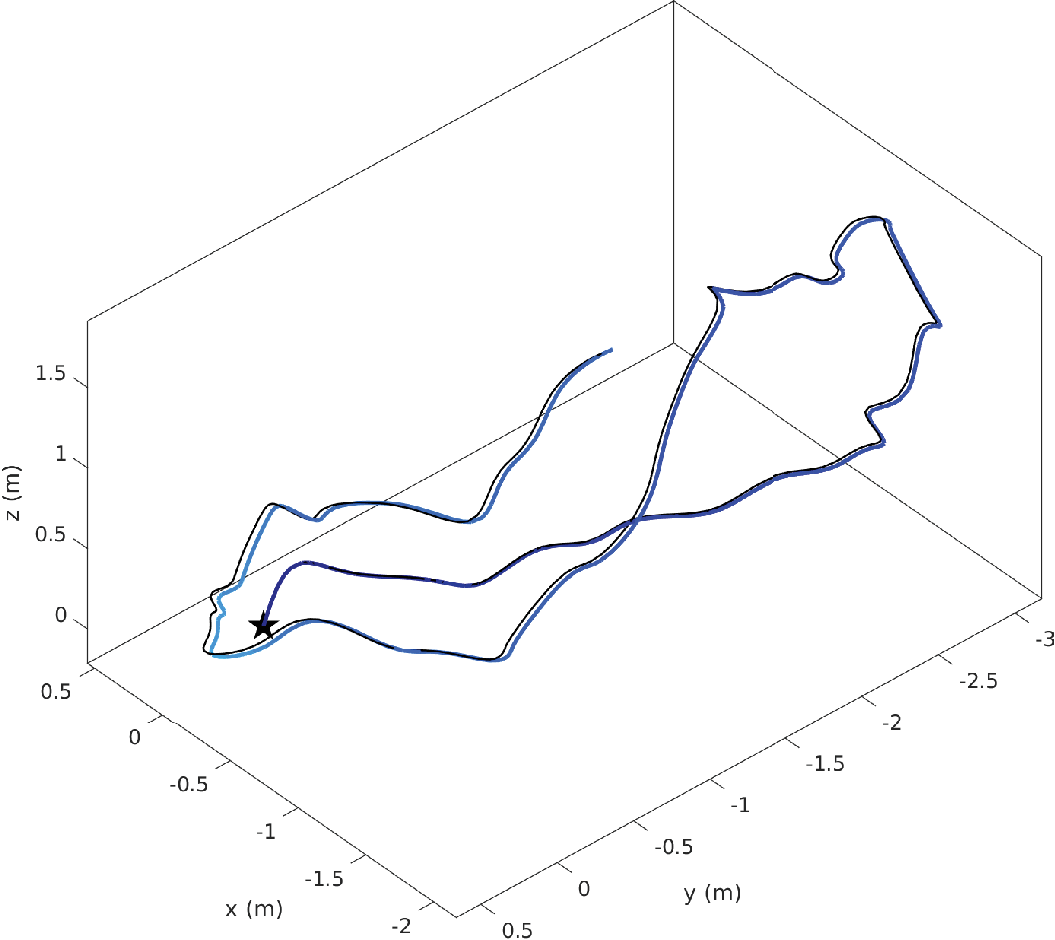}                   &
			\includegraphics[height=0.21\linewidth]{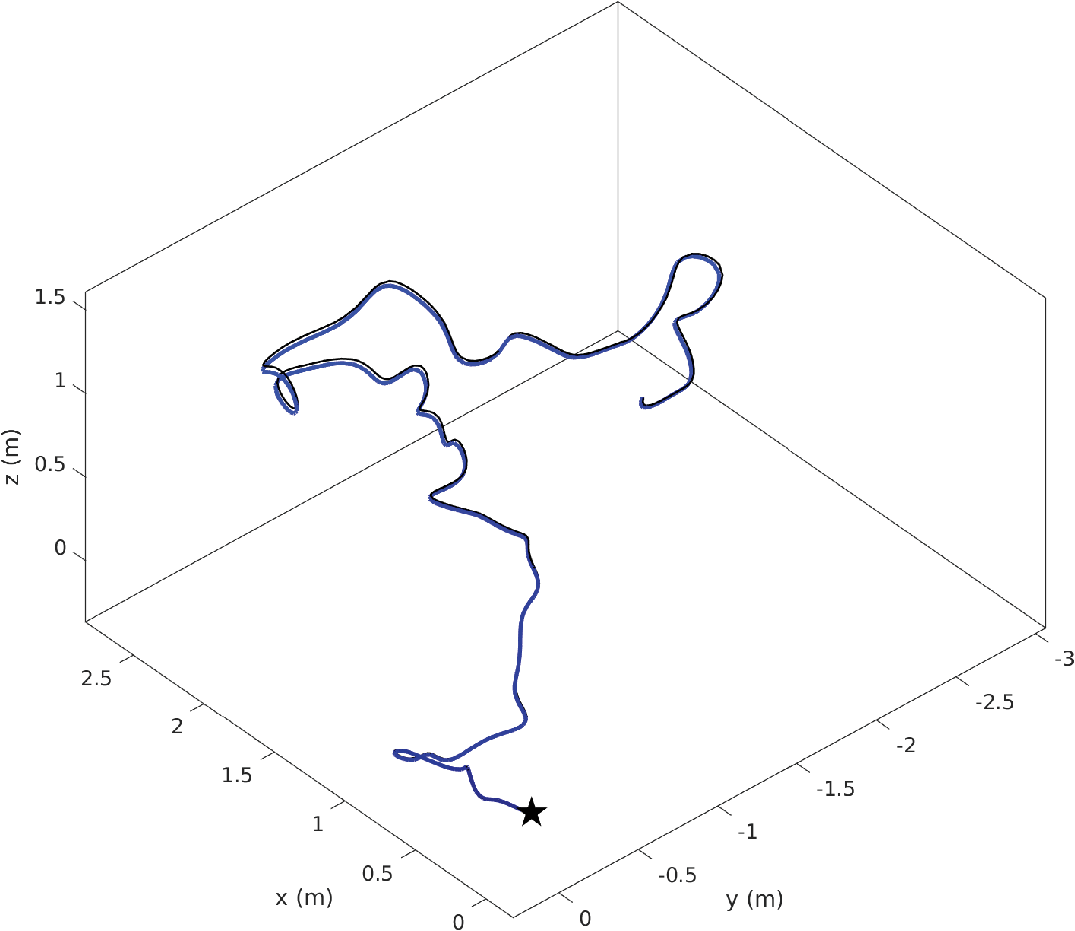}
			\\[0.3em]
		\end{tabular}
		\\
		 &
		\begin{tabular}{ccc}

			\includegraphics[width=0.18\linewidth]{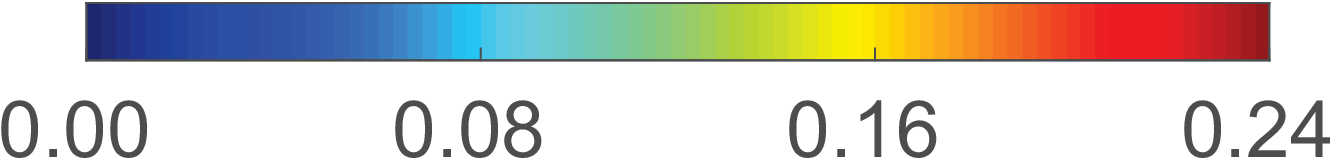} \quad \quad \quad \quad \quad &
			\includegraphics[width=0.18\linewidth]{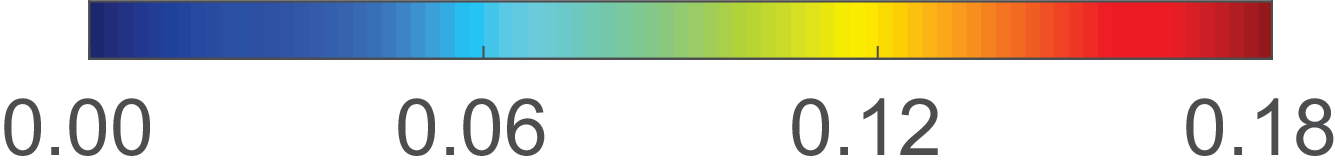}  \quad \quad \quad \quad      &
			\includegraphics[width=0.18\linewidth]{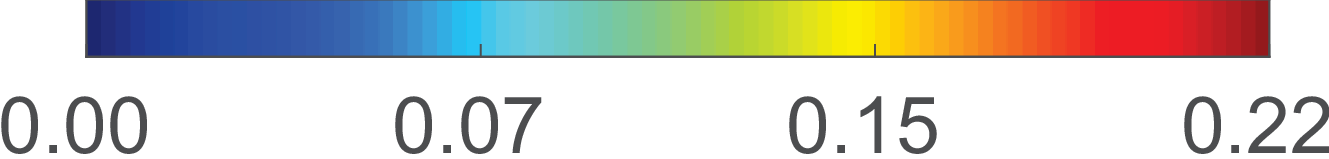}
		\end{tabular}
	\end{tabular}
	\\[-0.5em]
	\caption{Trajectories estimated by ORB-SLAM2~\cite{mur2017orb} and various PBA methods on three
		image sequences of our SpecularRooms dataset. The colored and black lines denote the estimated and ground truth trajectories, respectively.
		The color bar indicates the magnitude of the absolute trajectory error.}
	\label{fig_trajectories}
	\vspace{-0.9em}
\end{figure*}
\noindent\textbf{Point Encoder}. To extract multi-level geometry information of the scene, we adopt the point encoder developed based on the PointNet++~\cite{qi2017pointnet++}, following Synpt~\cite{synpt2020}. It is designed to extract information from points and their proximal surroundings. This model employs four layers of set abstraction and four levels of propagation, enabling the concurrent capture of localized and global features. The identified features are then interpolated within the feature propagation layers.

\noindent\textbf{Image Decoder}. Following~\cite{synpt2020}, point features are projected onto various levels of feature map spheres aligned with sub-pointset sizes. The image decoder employs these projected feature maps for generating the environment map. %

\noindent\textbf{RefineNet}. We use a RefineNet module to enhance coarse environment maps from the image decoder as~\cite{synpt2020}. In this module, we feed the color and depth information of points into the pipeline once again and introduce a 2D self-attention block to capture the long-range interactions.

\noindent\textbf{Loss Function}. We follow the setup of~\cite{lighthouse2020}, where the loss function comprises an adversarial loss term $L_\textnormal{adv}$ and a perceptual loss term. The perceptual loss quantifies the high-level discrepancies between images. Following ~\cite{lighthouse2020}, we use a pre-trained VGG-19 network
for feature extraction
and compute the $L_1$ loss of features across layers as ${L}_\textnormal{vgg}$. The total loss function ${L_\textnormal{light}}$ is defined as a weighted sum:
\begin{equation}
	{L_\textnormal{light}}=\lambda_{\textnormal{vgg}}{L}_{\textnormal{vgg}}+\lambda_{\textnormal{adv}}{L}_{\textnormal{adv}},
\end{equation}
where $\lambda_{\textnormal{vgg}}$ and $\lambda_{\textnormal{adv}}$ are the weights of perceptual loss and adversarial loss.

\section{EXPERIMENTS}
We first introduce our SpecularRooms dataset. Then we compare our PBA, material estimation, and illumination methods with corresponding state-of-the-art works. Ablation studies, additional results, and more implementation details are available in our supplementary document. All the following tests were conducted on a computer equipped with NVIDIA RTX A4000 GPU.

\begin{figure*}[htbp]
	\footnotesize
	\centering
	\renewcommand{\tabcolsep}{0.5pt}
	\begin{tabular}{c|c|c|c|c}
		\parbox{2cm}{\textsf{Input Image}}~
		 &
		\begin{tabular}{c|c|c|c|c|c}

			\\
			\includegraphics[width=0.12\linewidth]{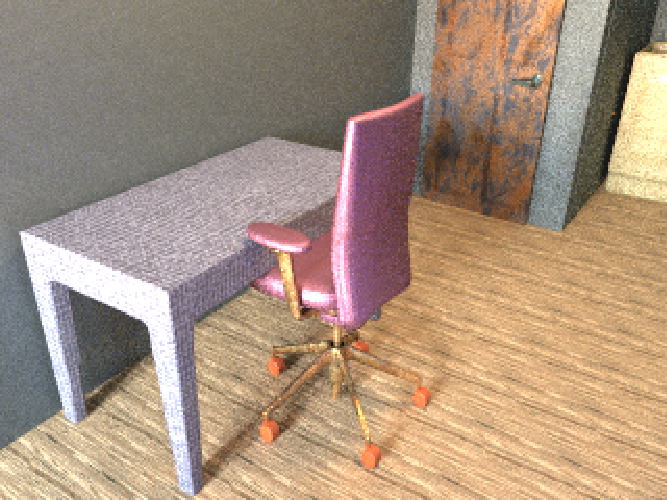}  &
			\includegraphics[width=0.12\linewidth]{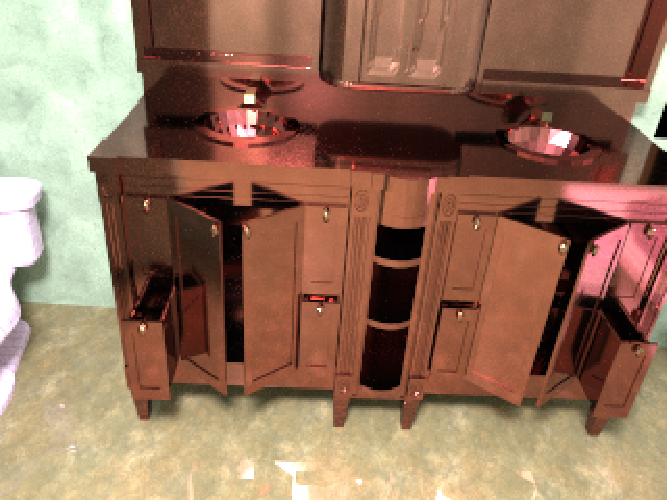}  &
			\includegraphics[width=0.12\linewidth]{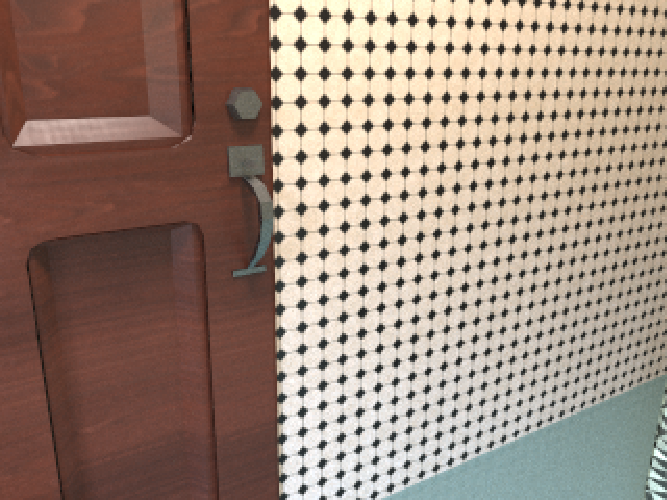} &
			\includegraphics[width=0.12\linewidth]{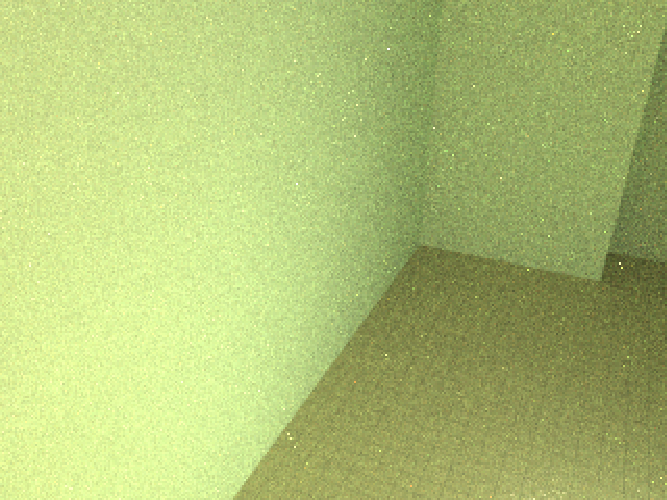} &
			\includegraphics[width=0.12\linewidth]{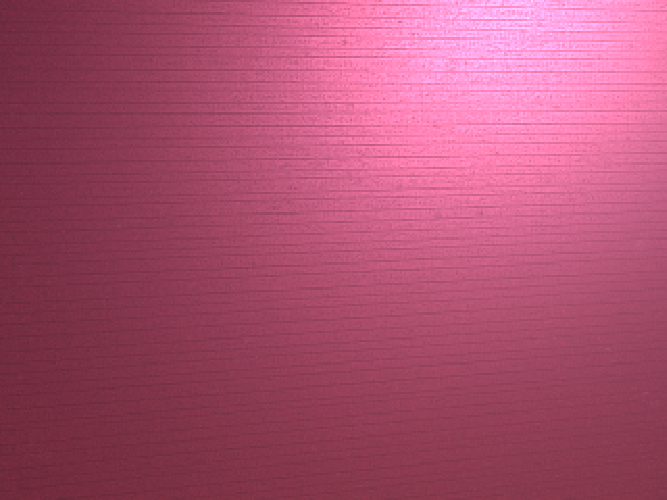} &
			\includegraphics[width=0.12\linewidth]{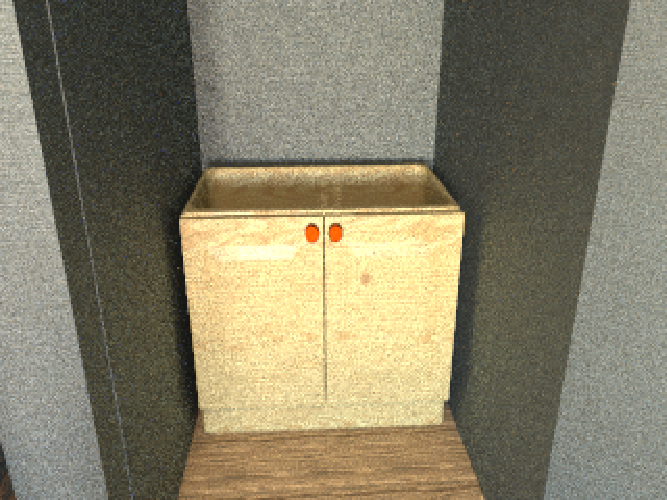}
			\\[0.3em]
		\end{tabular}
		\\
		\parbox{2cm}{\textsf{IRCIS~\cite{li2020inverse}}}

		 &
		\begin{tabular}{c|c|c|c|c|c}
			\includegraphics[width=0.12\linewidth]{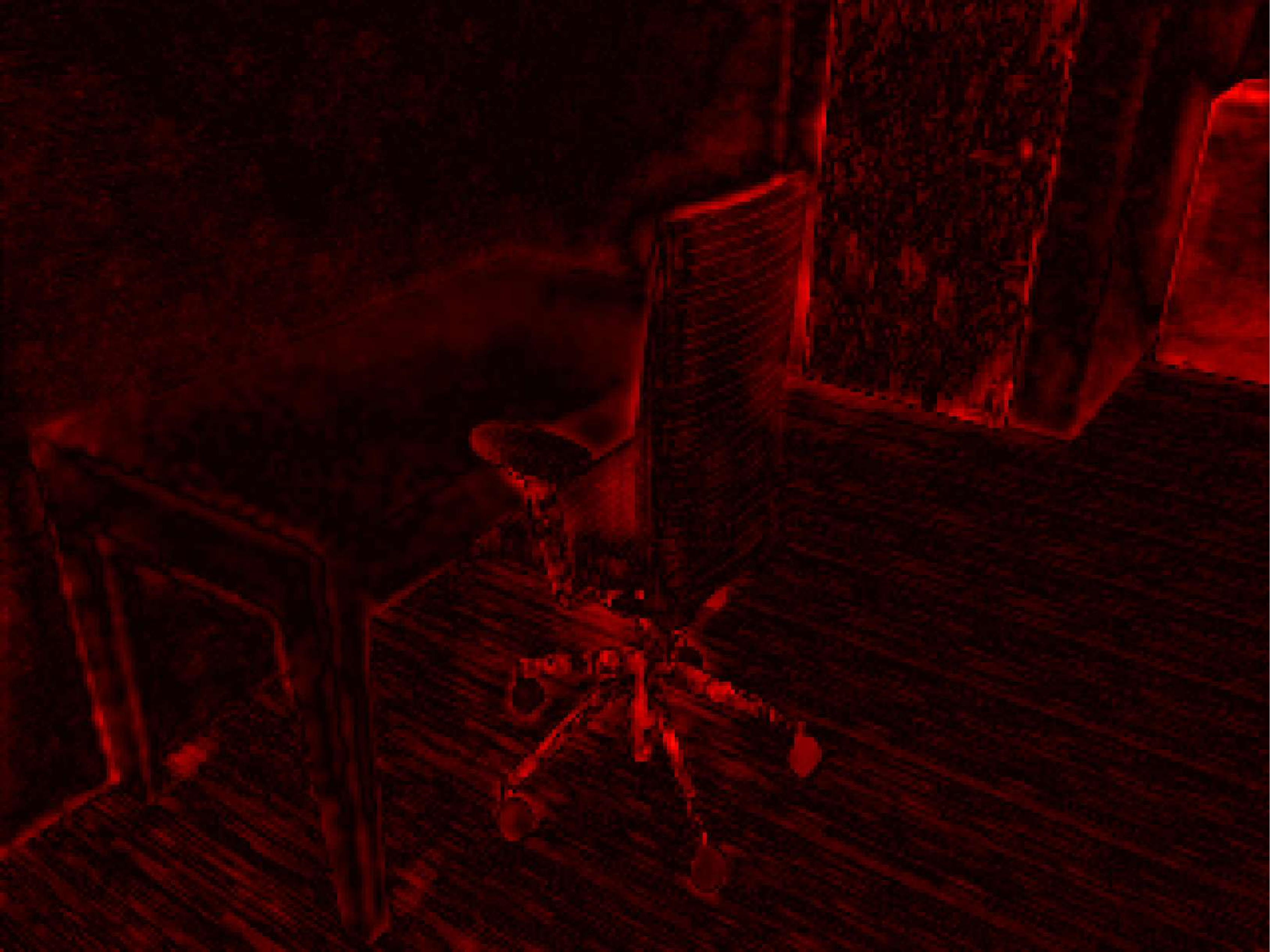}  &
			\includegraphics[width=0.12\linewidth]{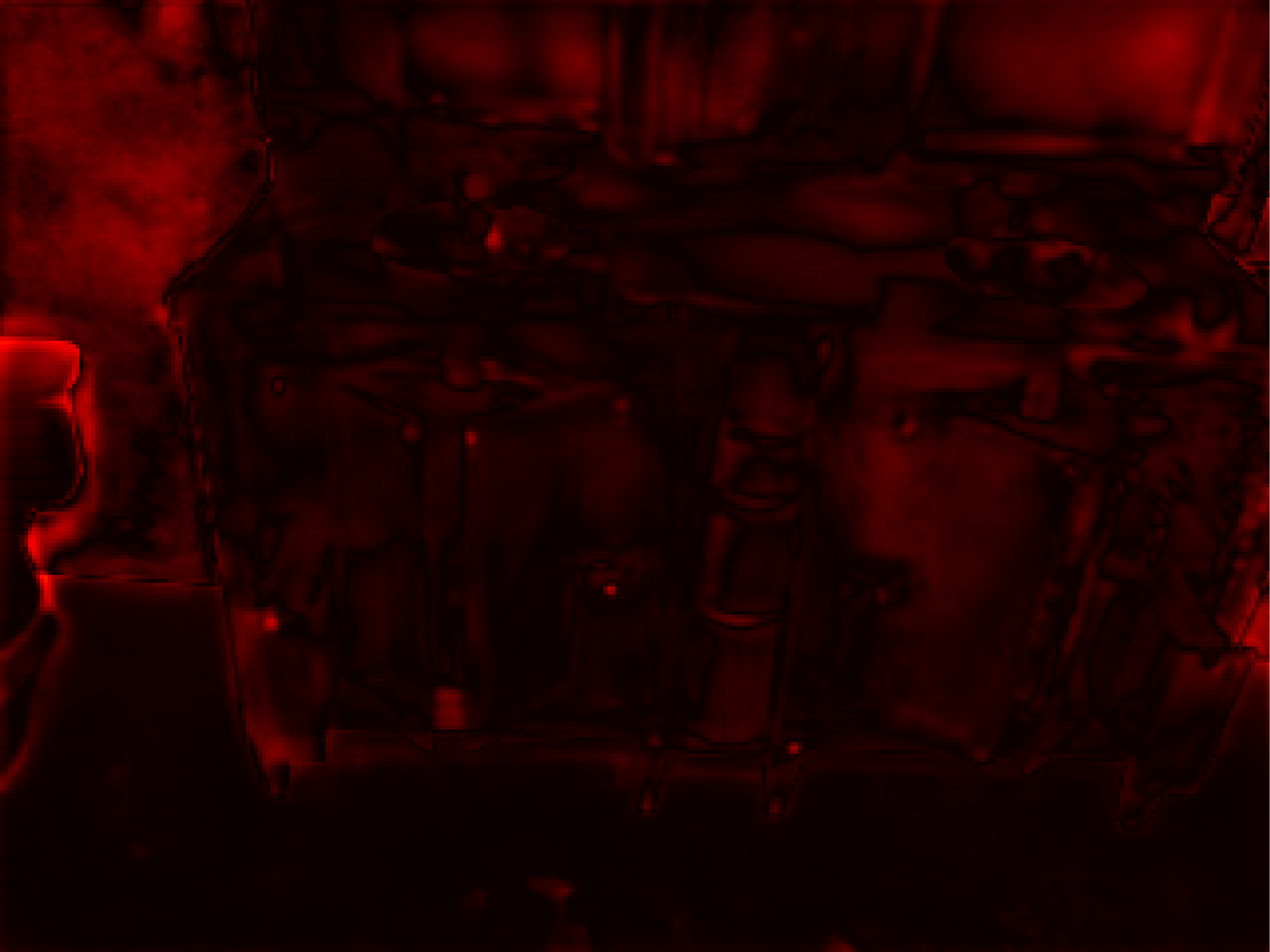}  &
			\includegraphics[width=0.12\linewidth]{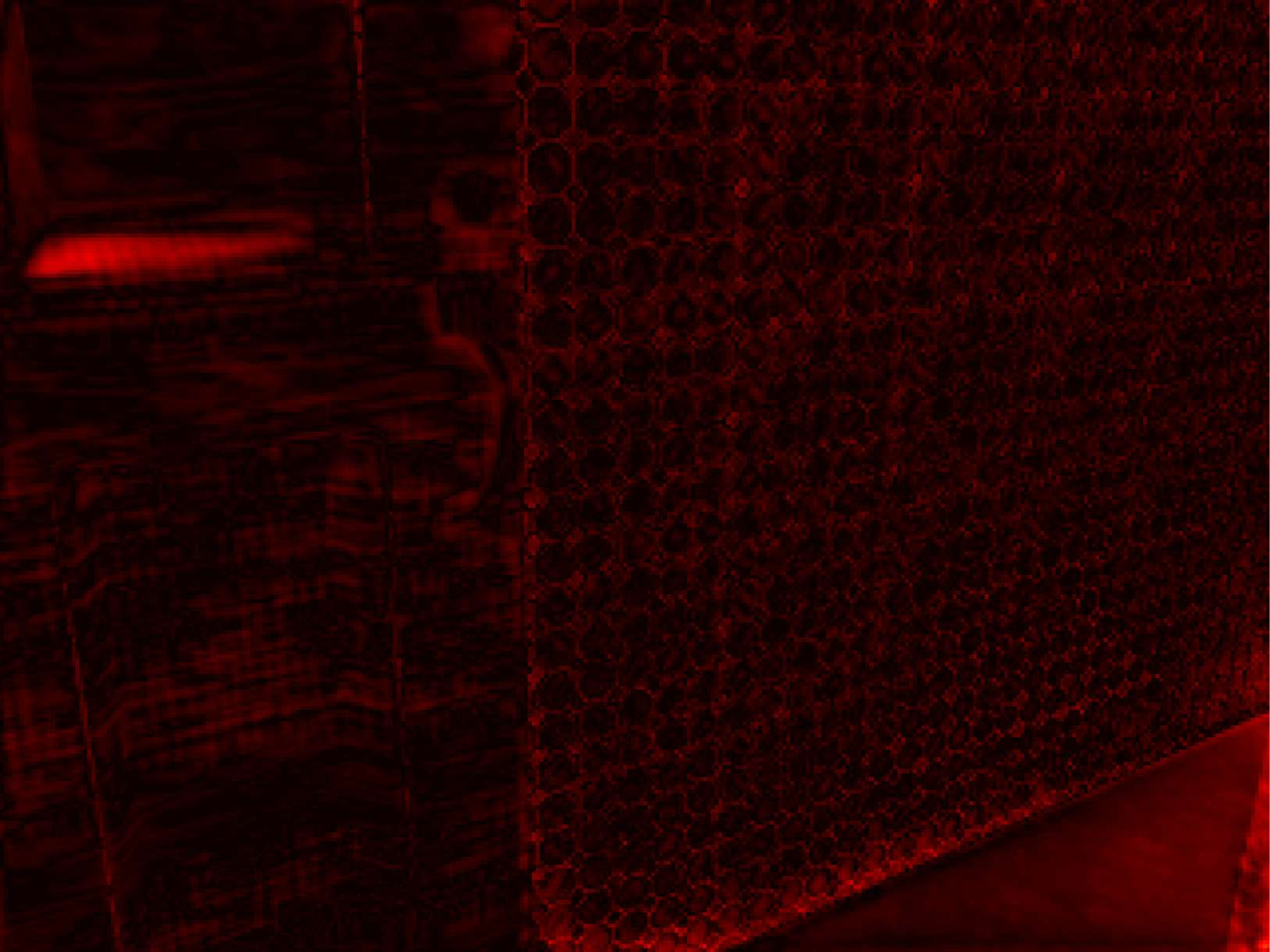} &
			\includegraphics[width=0.12\linewidth]{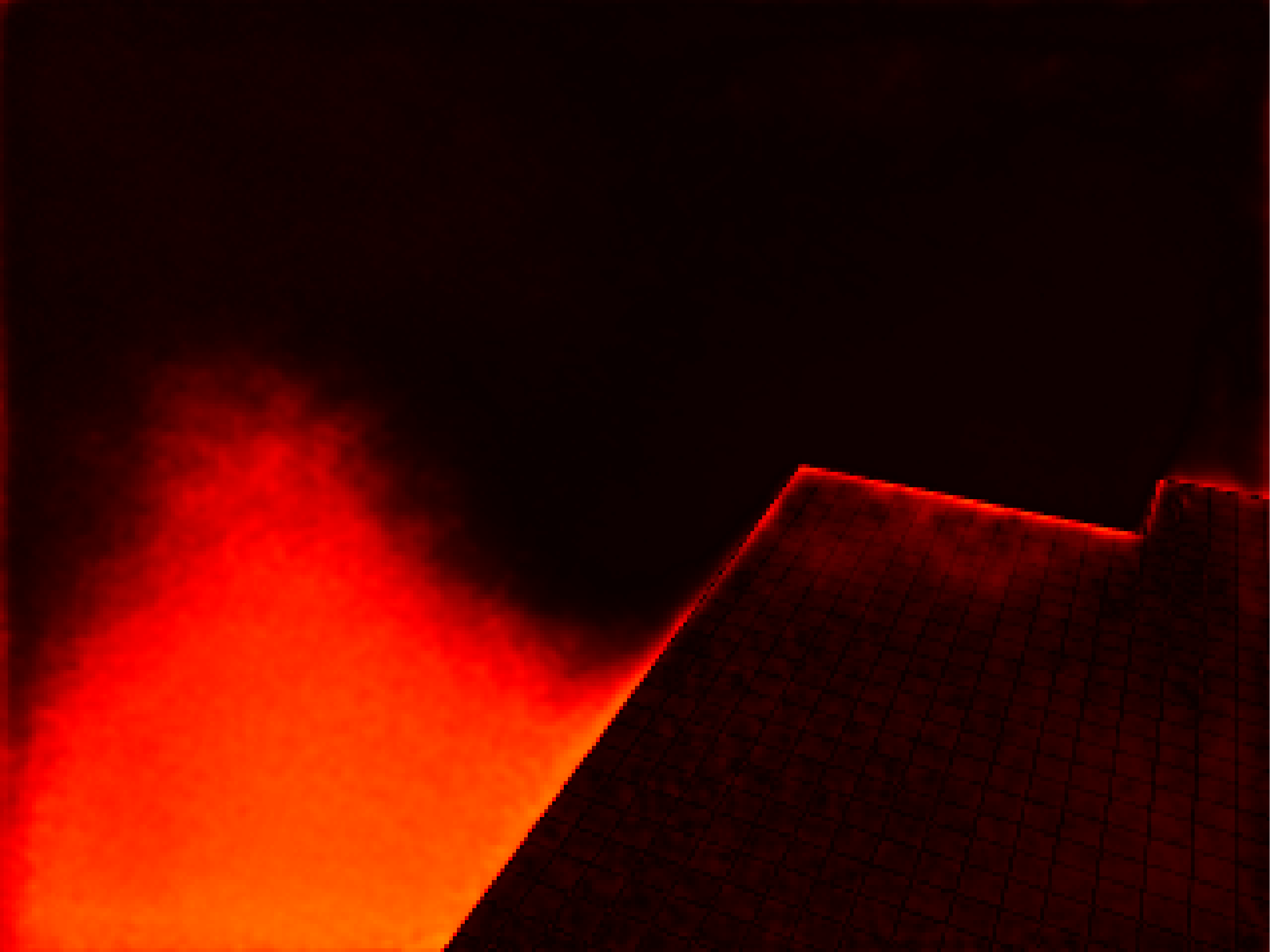} &
			\includegraphics[width=0.12\linewidth]{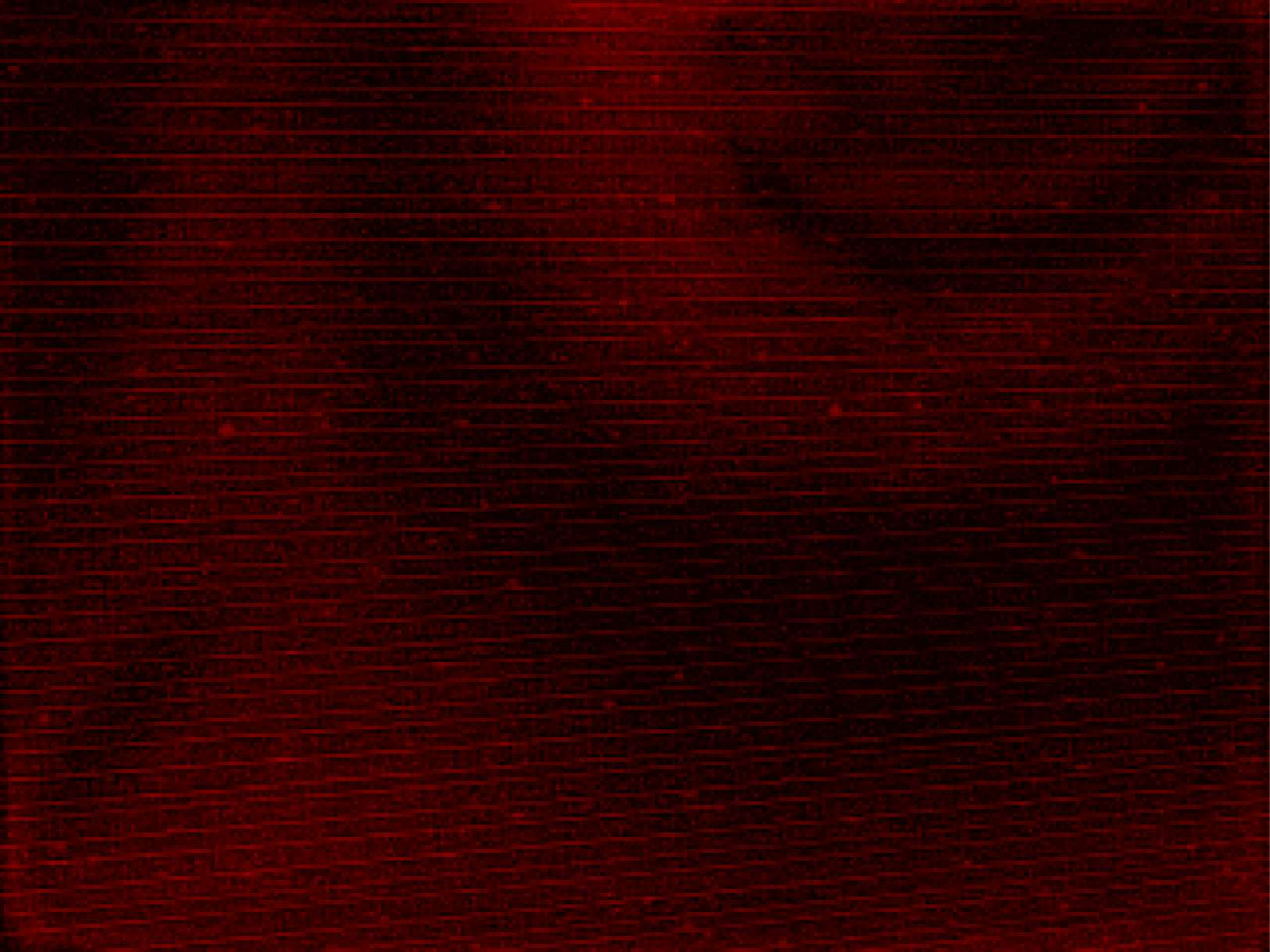} &
			\includegraphics[width=0.12\linewidth]{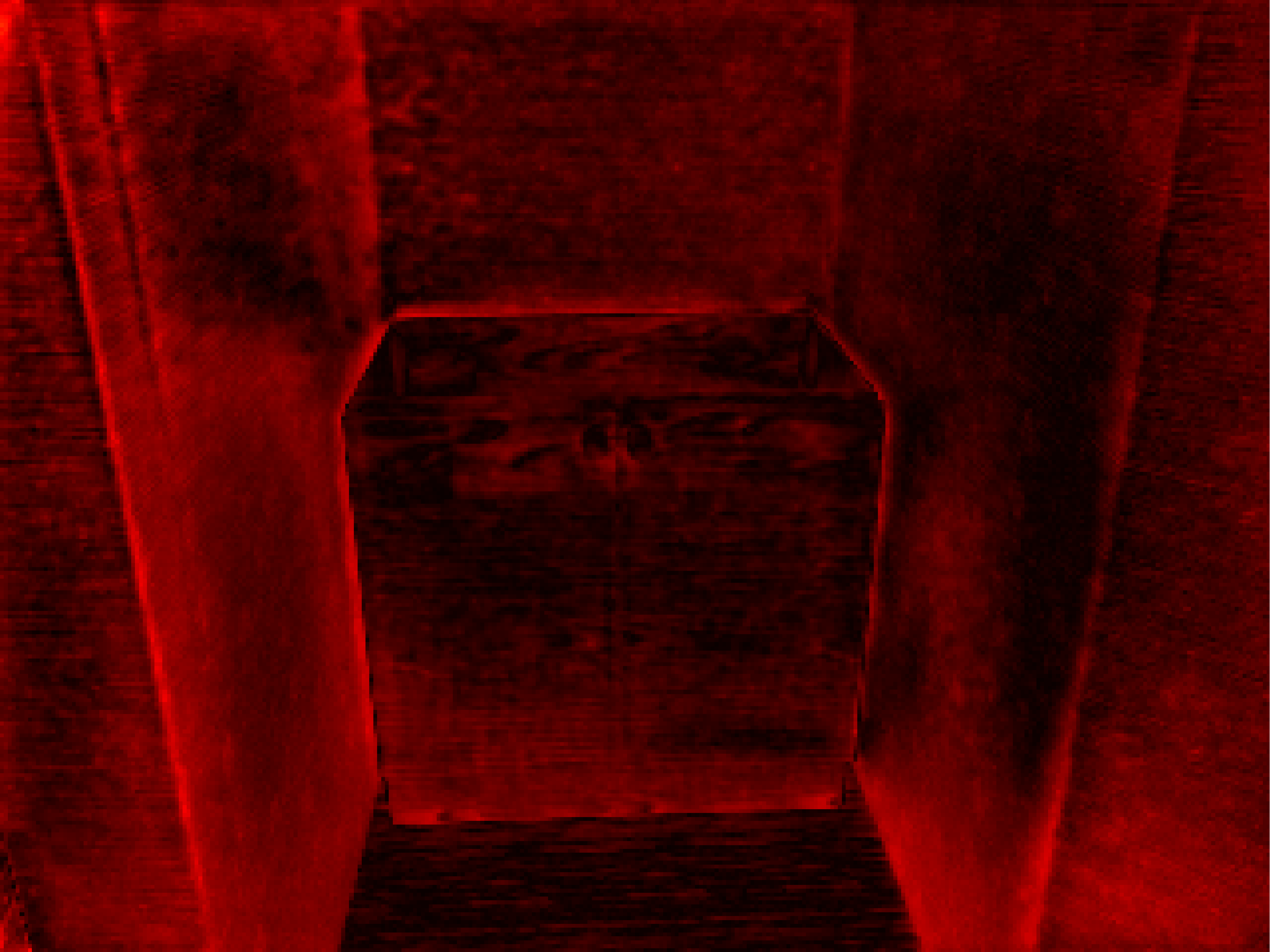}
			\\[0.1em]
		\end{tabular}

		\\
		\parbox{2cm}{\textsf{Ours}}
		 &
		\begin{tabular}{c|c|c|c|c|c}
			\includegraphics[width=0.12\linewidth]{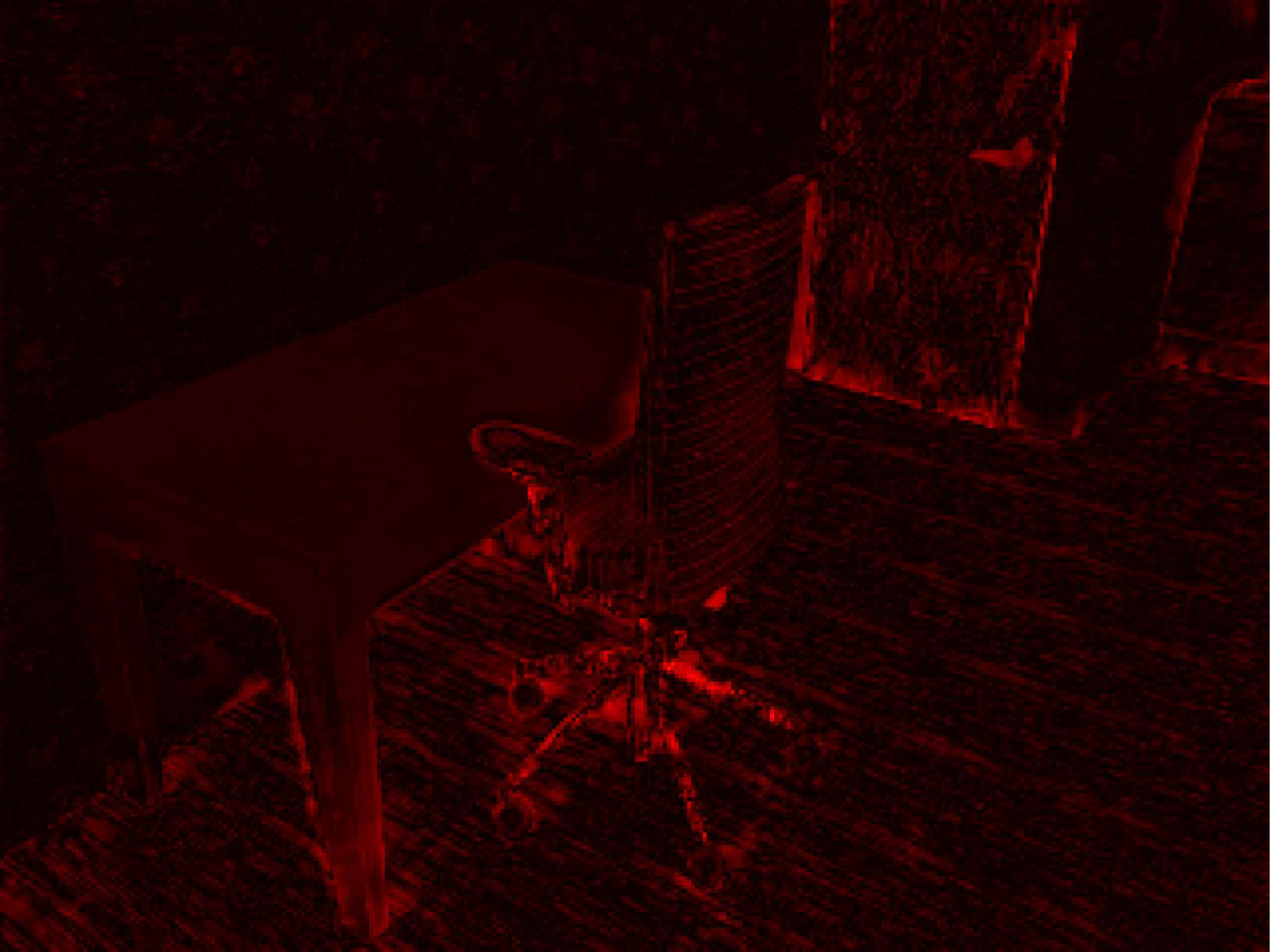} &
			\includegraphics[width=0.12\linewidth]{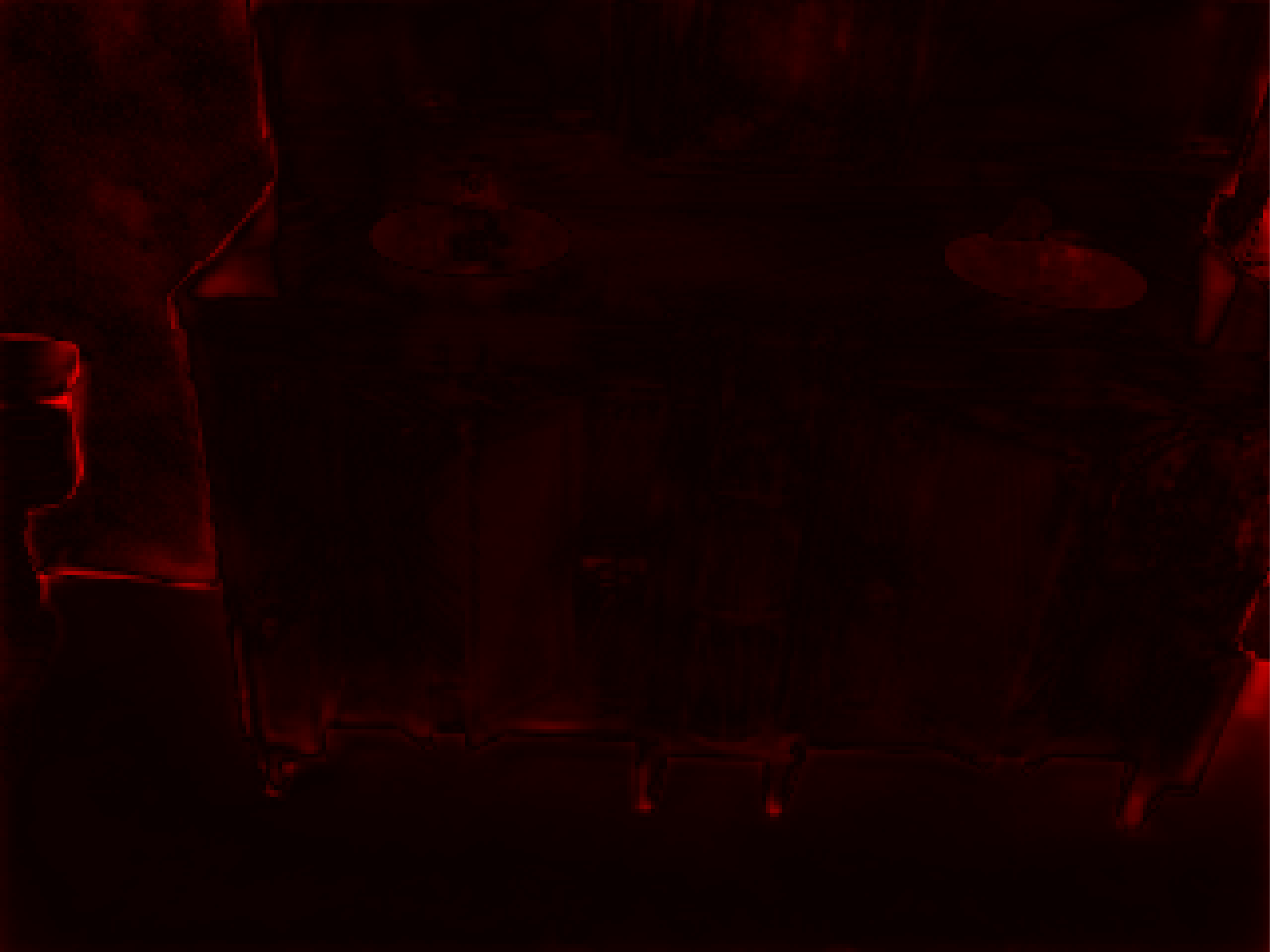}  &
			\includegraphics[width=0.12\linewidth]{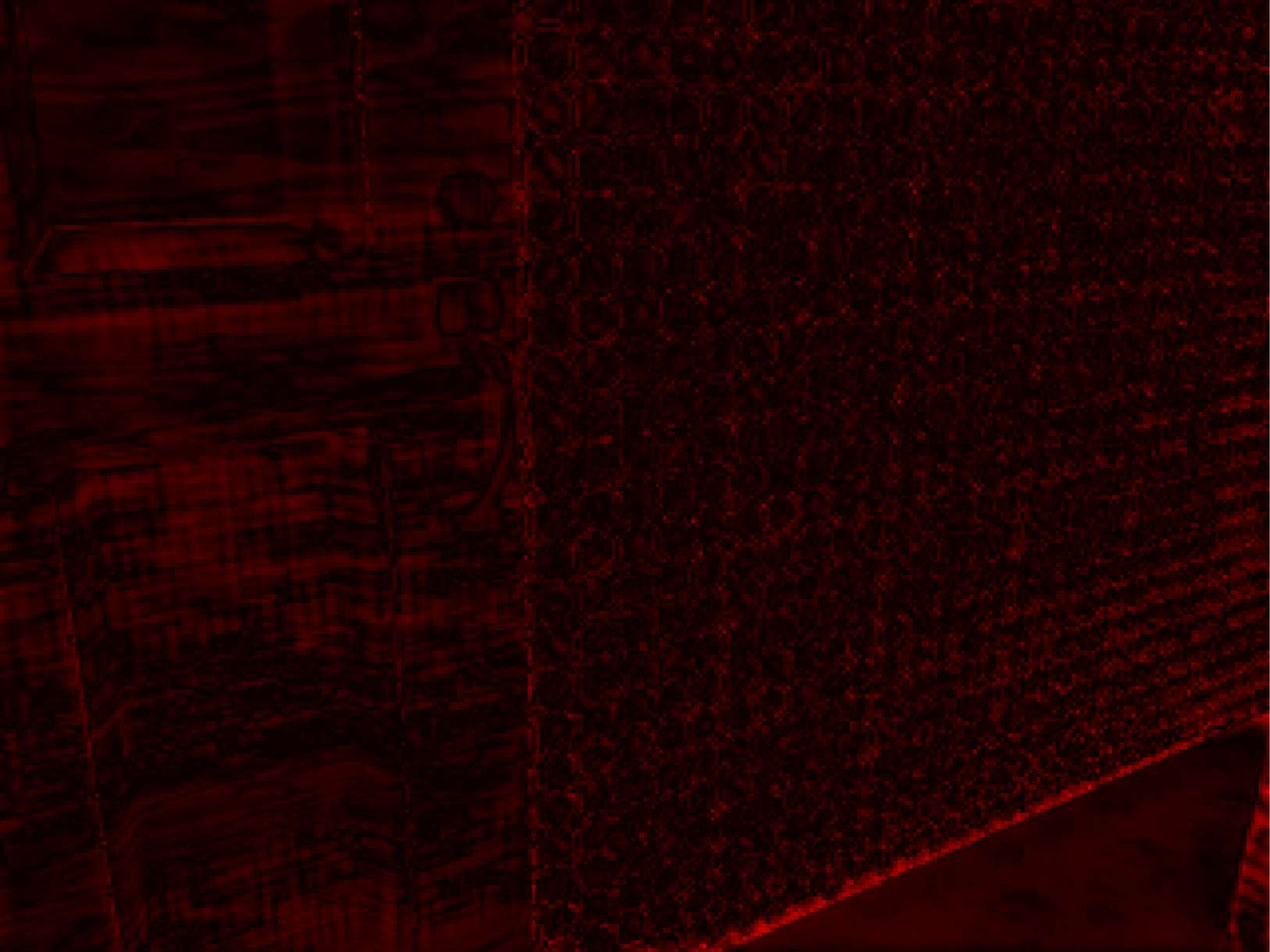} &
			\includegraphics[width=0.12\linewidth]{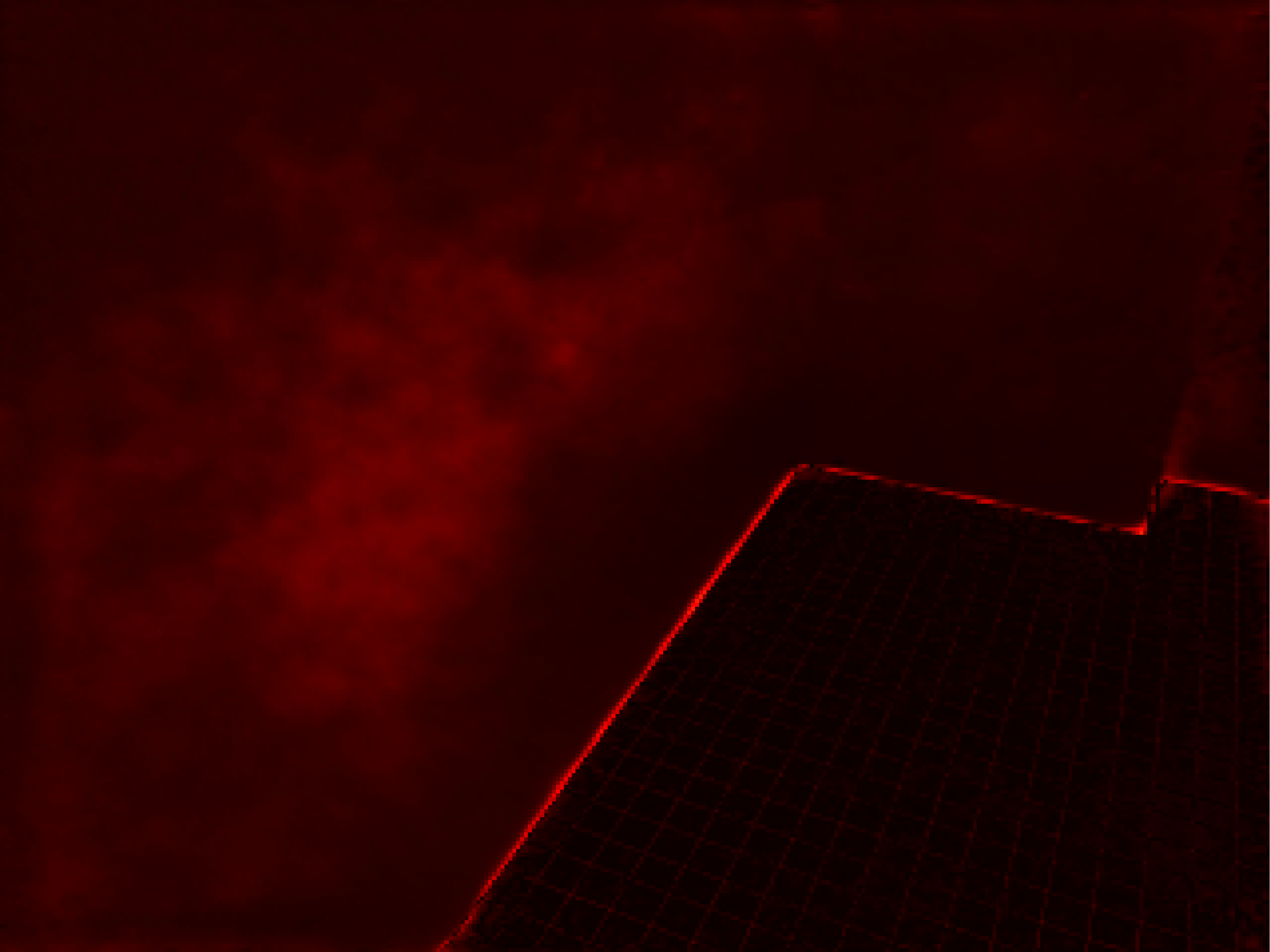} &
			\includegraphics[width=0.12\linewidth]{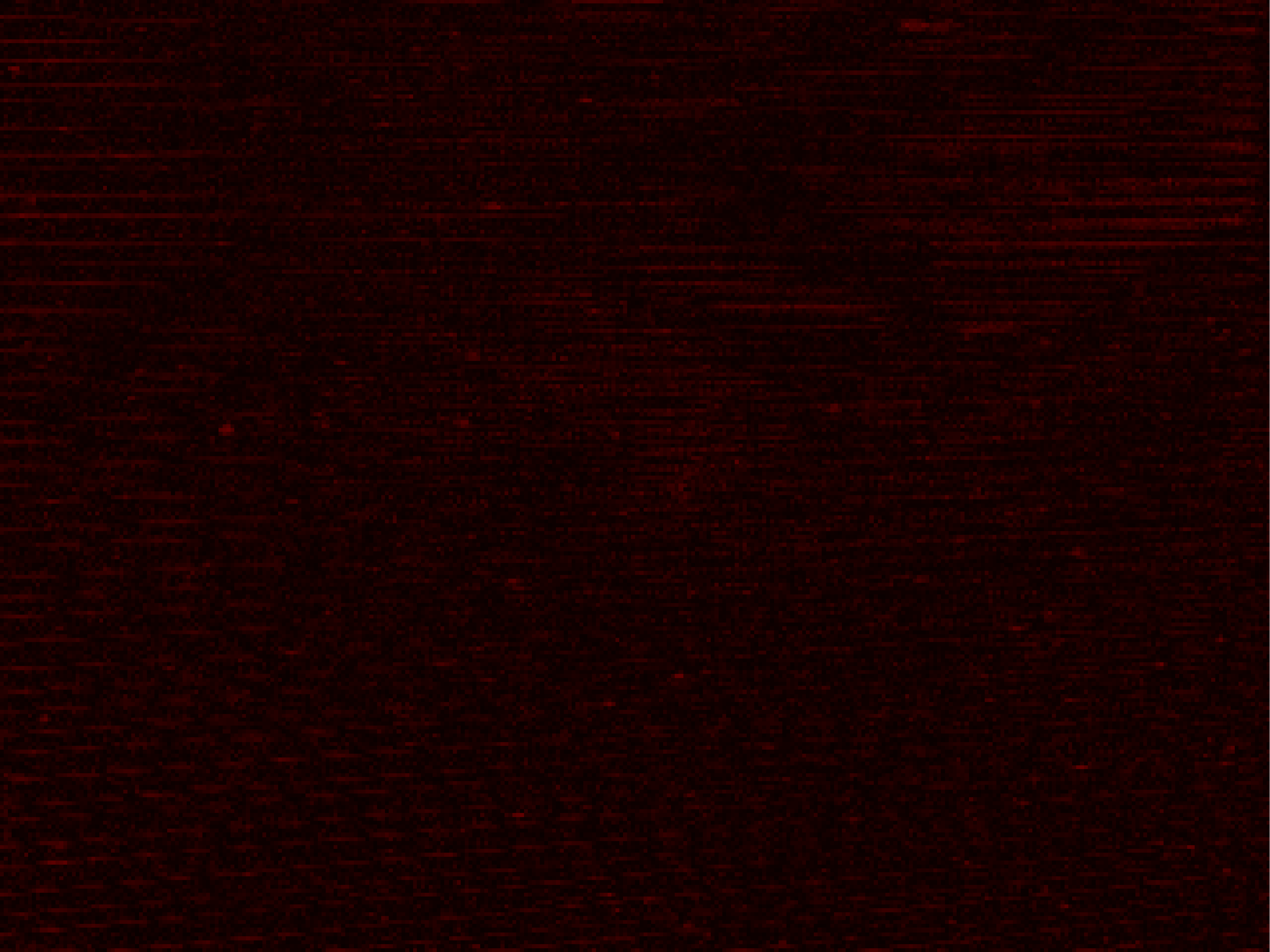} &
			\includegraphics[width=0.12\linewidth]{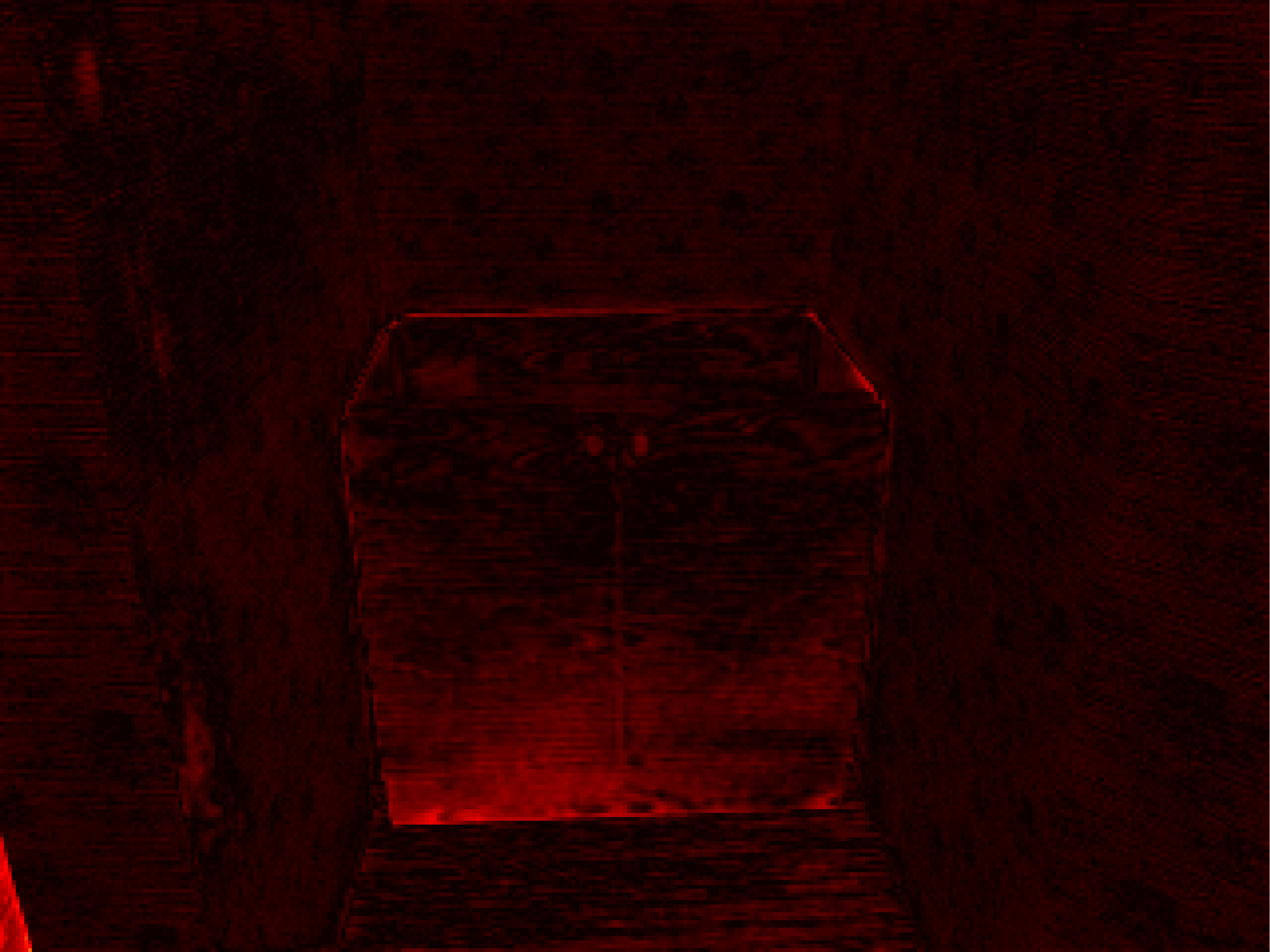}
			\\[0.1em]
		\end{tabular}
		\\
		\parbox{2cm}{\textsf{}}
		 &
		\begin{tabular}{c}
			\includegraphics[width=0.3\linewidth]{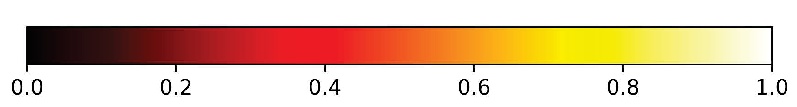}
		\end{tabular}
	\end{tabular}
	\caption{Error maps of roughness estimation results on our SpecularRooms dataset.
		Here we list input images (in the first row) and the corresponding error maps of roughness predictions. The second row corresponds to \textsf{IRCIS} and the third row corresponds to our method. The color bar indicates the magnitude of the difference between roughness predictions and ground truth. Lighter pixels correspond to larger errors. Note that our method works better on glossy and featureless surfaces. %
	}
	\label{fig_material}
\end{figure*}

\begin{figure*}[t]
	\footnotesize
	\centering
	\renewcommand{\tabcolsep}{0.4pt}
	\renewcommand\arraystretch{1.1} %
	\begin{tabular}{c|c}
		\textsf{Input}~
		 &
		\begin{tabular}{c|c|c|c}
			\textit{Meeting Room}                                                  & \textit{Apartment} & \textit{Office} & \textit{Club}
			\\
			\includegraphics[width=0.17\linewidth]{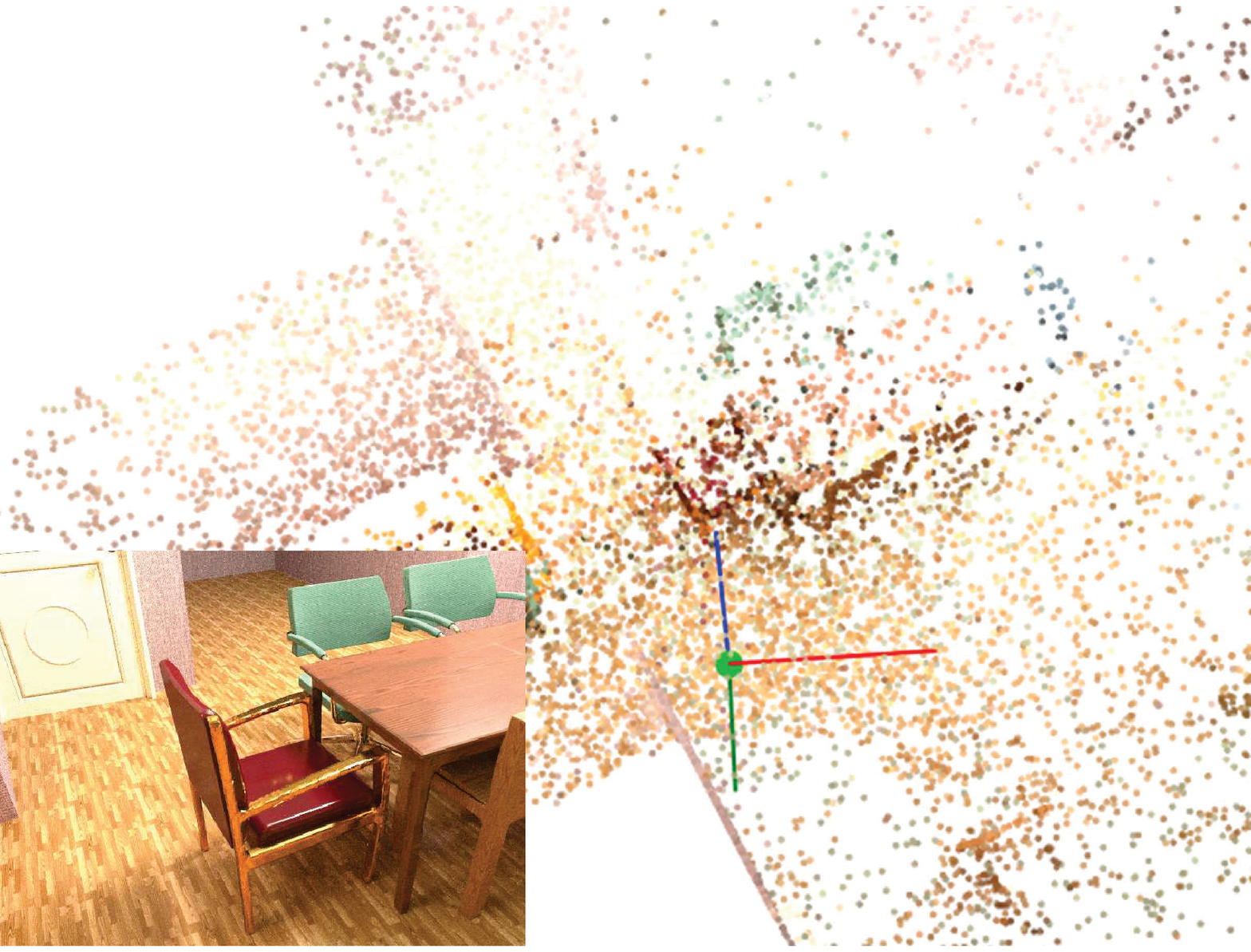} &
			\includegraphics[width=0.17\linewidth]{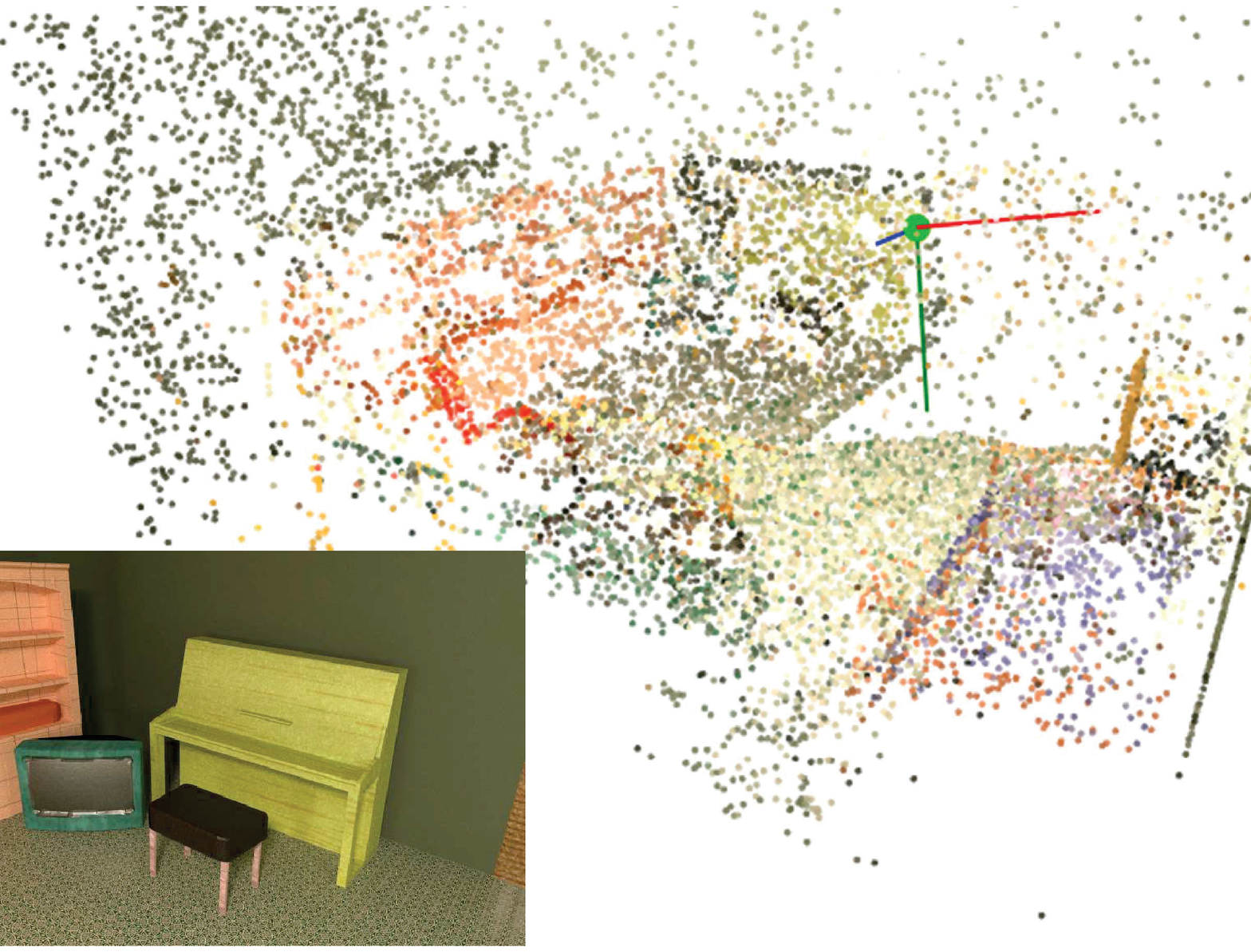} &
			\includegraphics[width=0.17\linewidth]{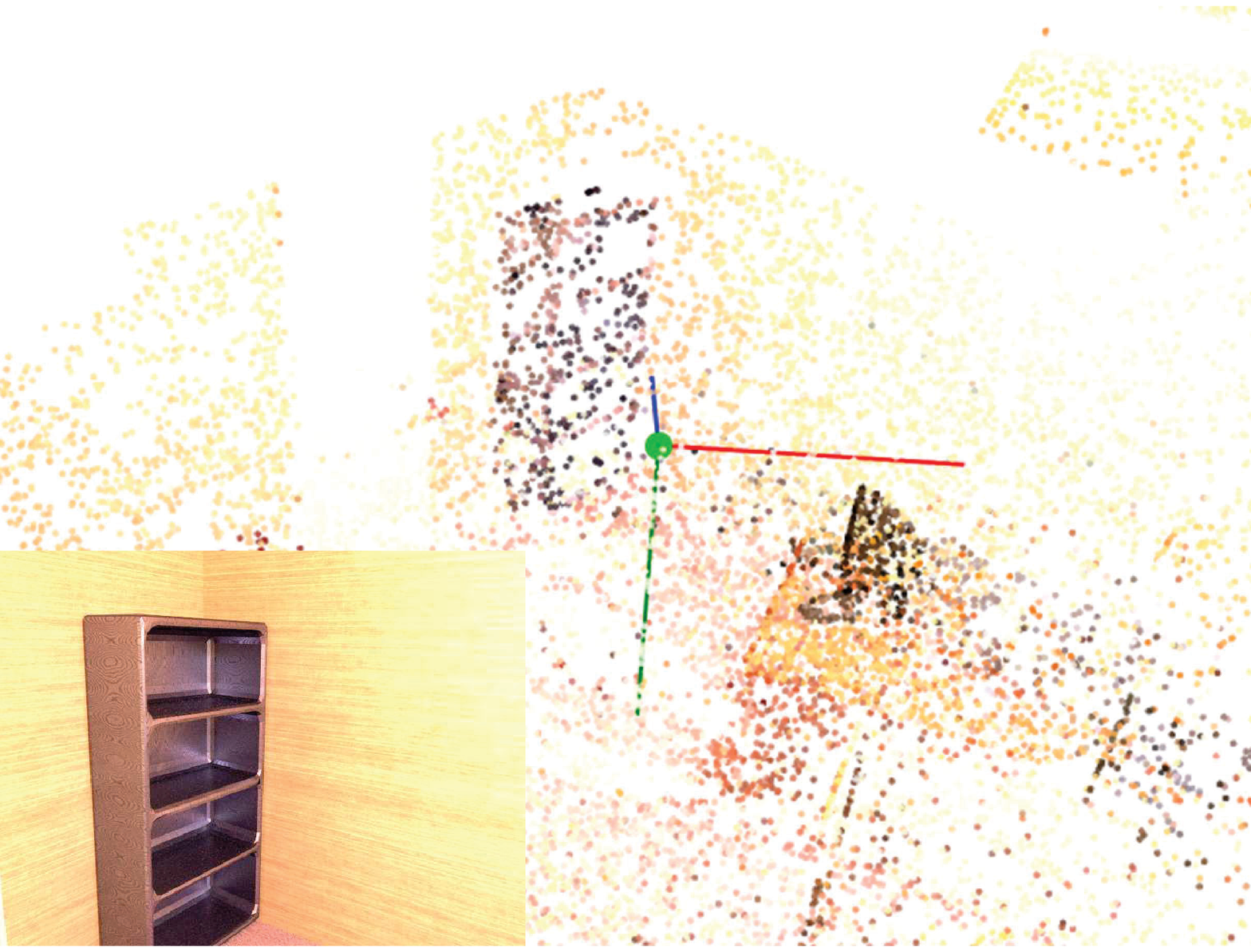} &
			\includegraphics[width=0.17\linewidth]{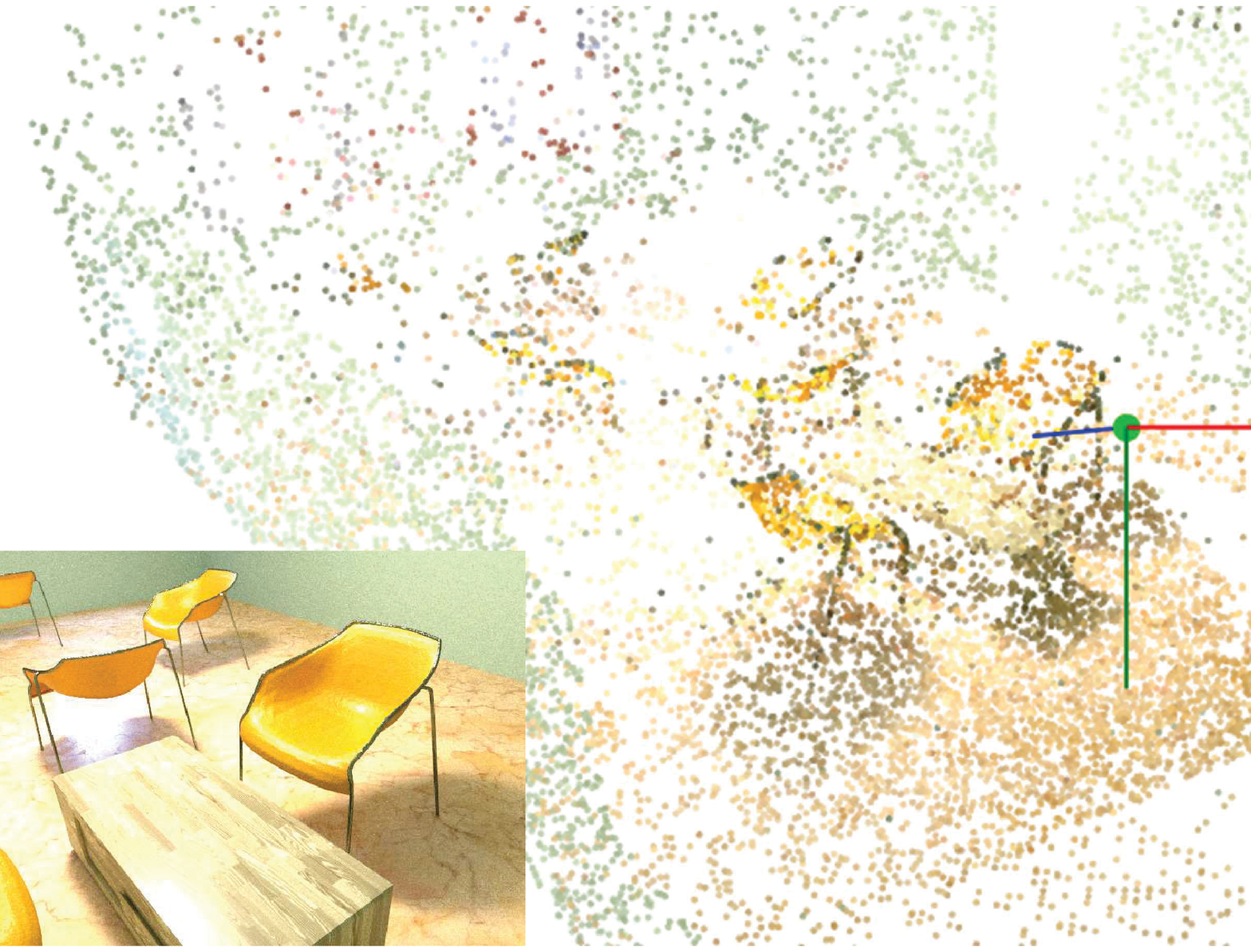}
			\\[0.3em]
		\end{tabular}
		\\
		\textsf{Lighthouse~\cite{lighthouse2020}}~
		 &
		\begin{tabular}{c|c|c|c}
			\includegraphics[width=0.17\linewidth]{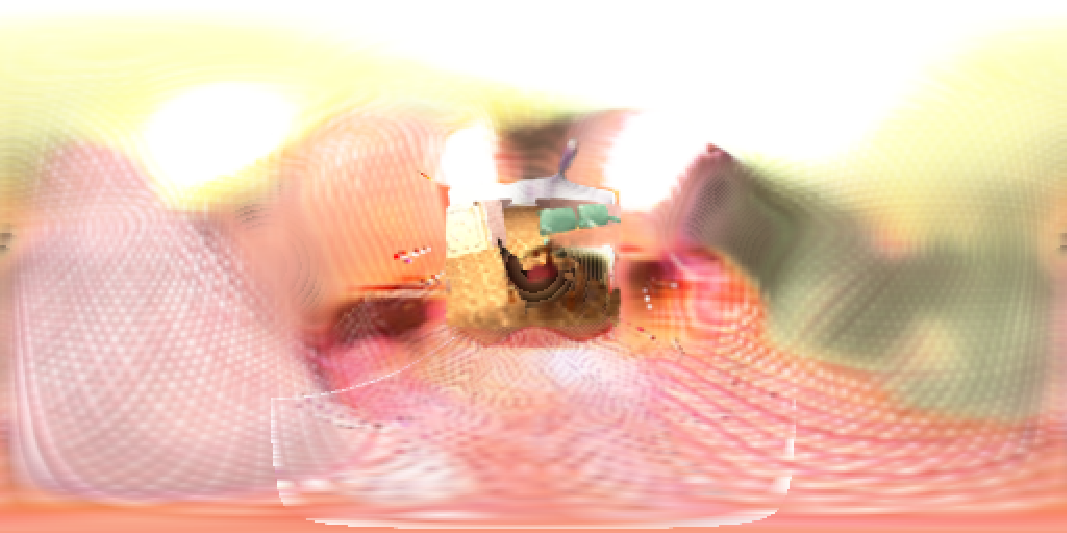}  &
			\includegraphics[width=0.17\linewidth]{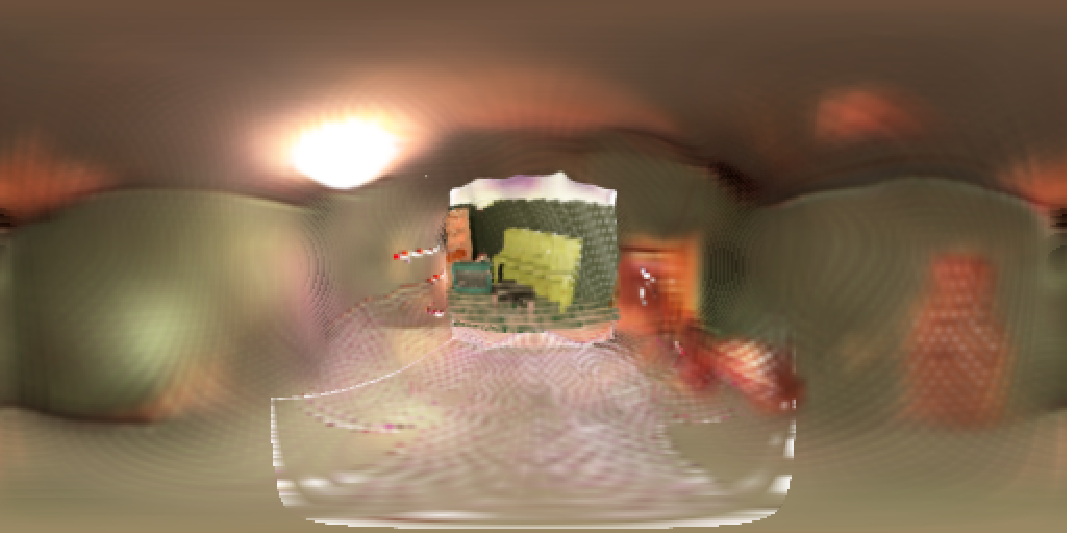}  &
			\includegraphics[width=0.17\linewidth]{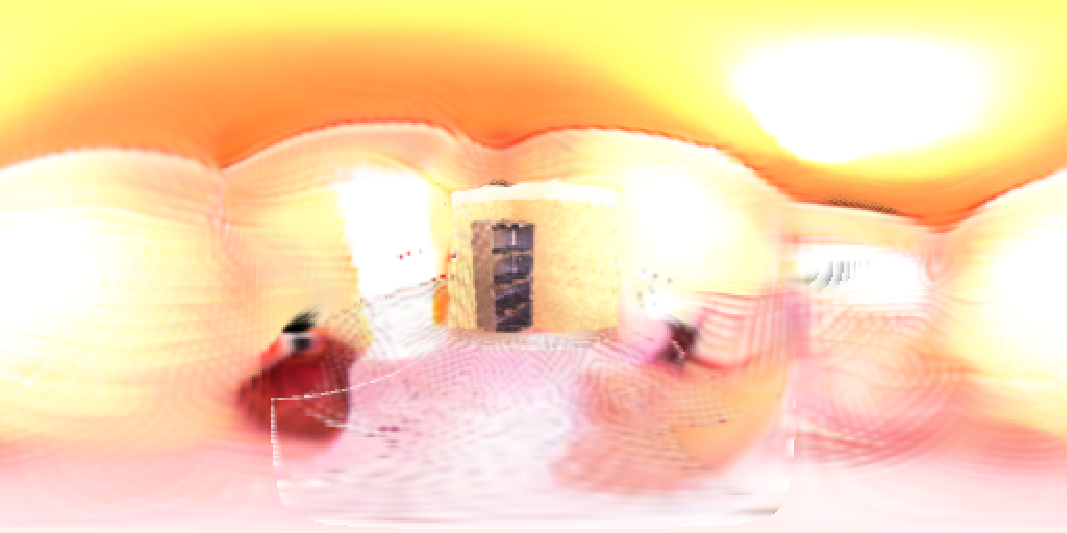} &
			\includegraphics[width=0.17\linewidth]{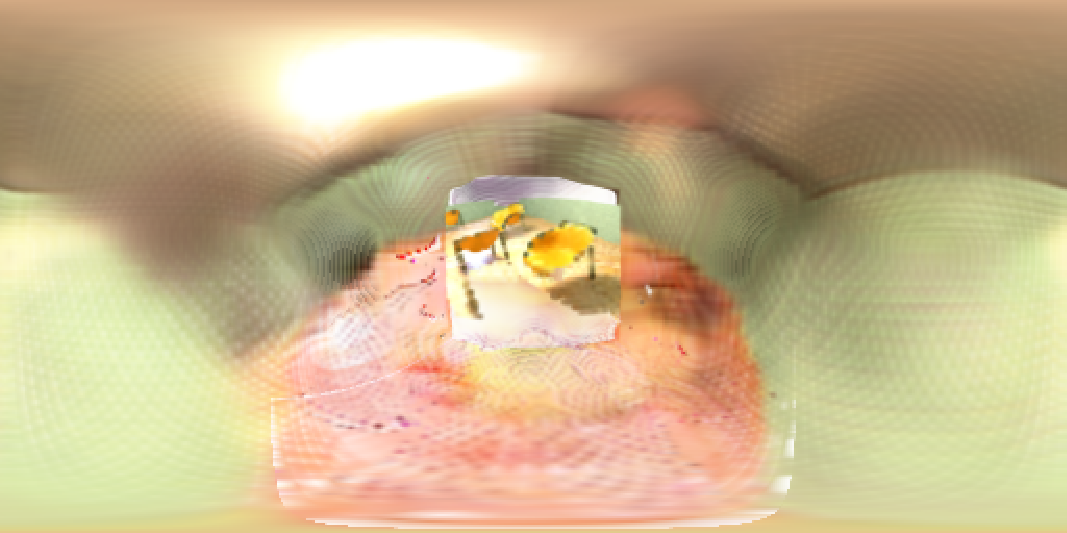}
			\\[0.3em]
		\end{tabular}
		\\
		\textsf{Ours}
		 &
		\begin{tabular}{c|c|c|c}
			\includegraphics[width=0.17\linewidth]{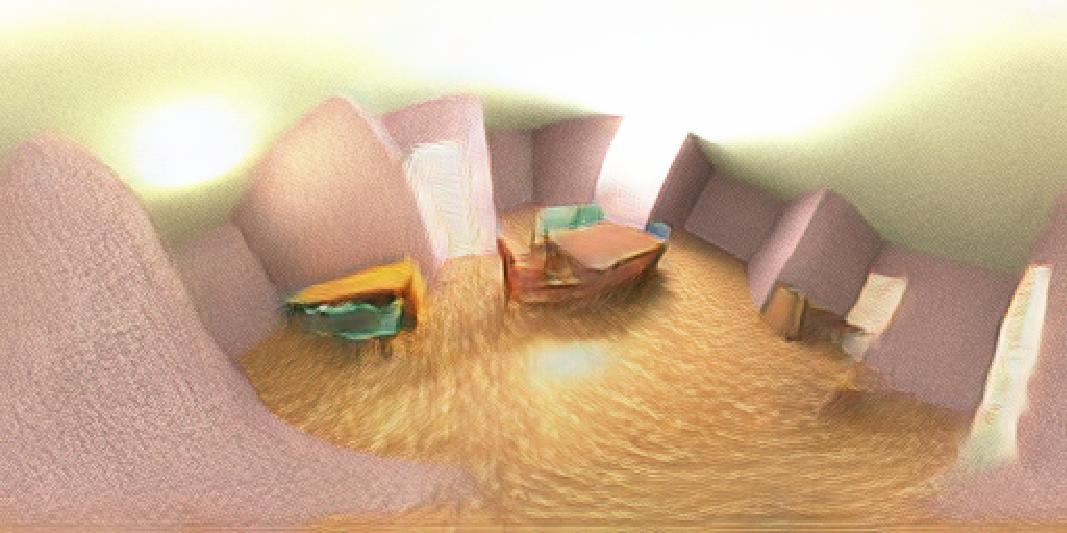} &
			\includegraphics[width=0.17\linewidth]{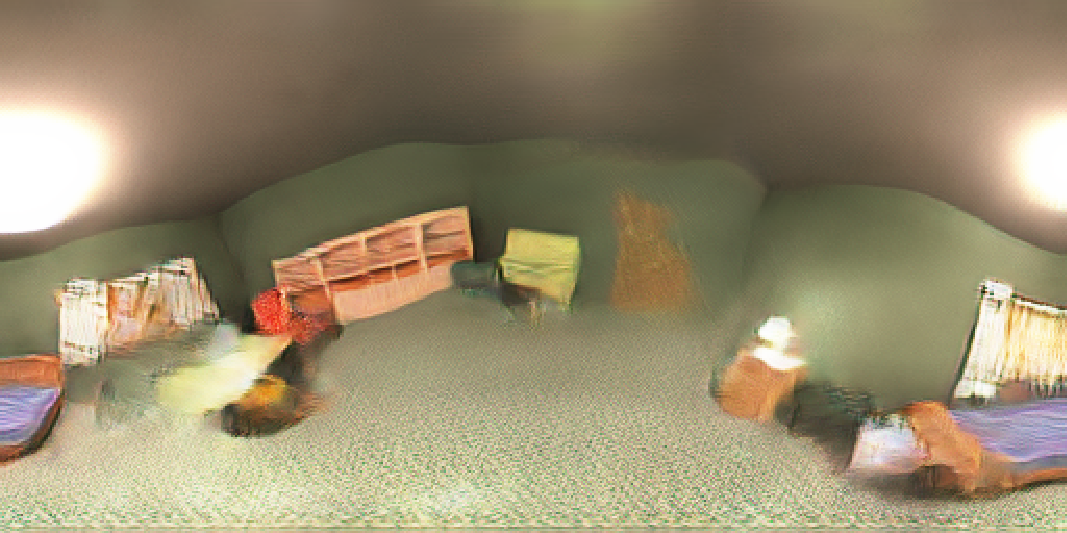} &
			\includegraphics[width=0.17\linewidth]{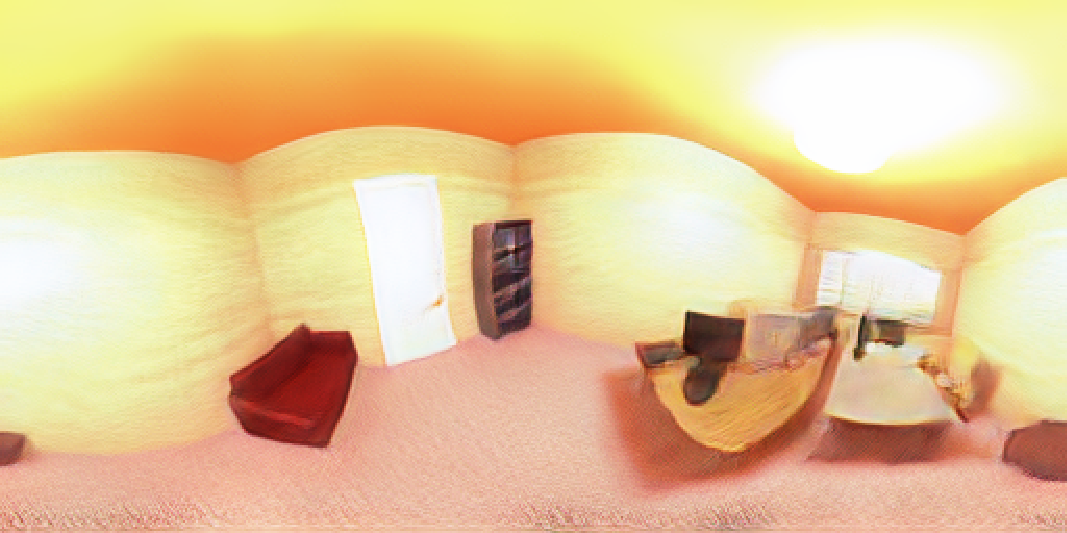} &
			\includegraphics[width=0.17\linewidth]{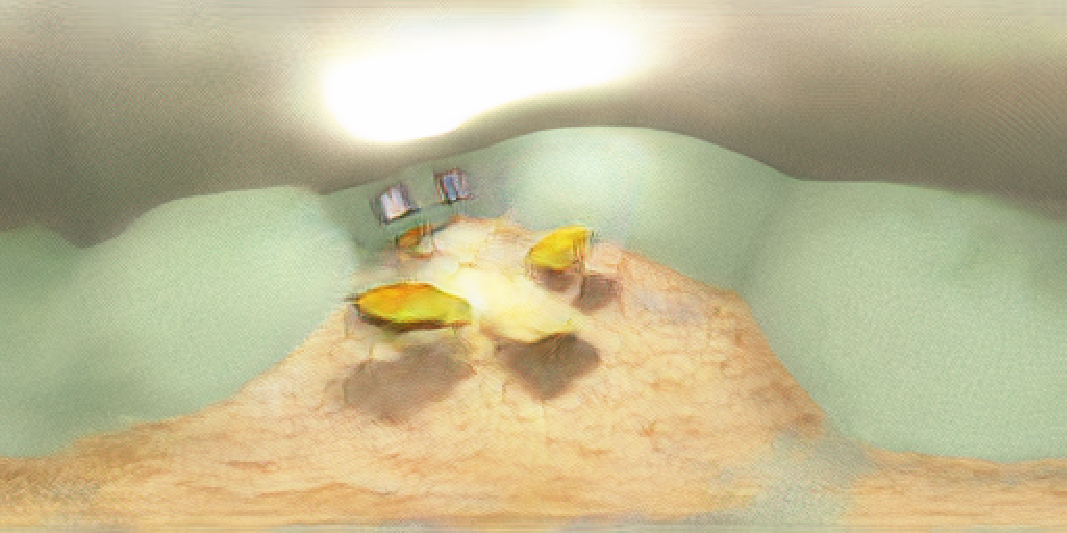}
			\\[0.3em]
		\end{tabular}
		\\
		\textsf{Ground Truth}
		 &
		\begin{tabular}{c|c|c|c}
			\includegraphics[width=0.17\linewidth]{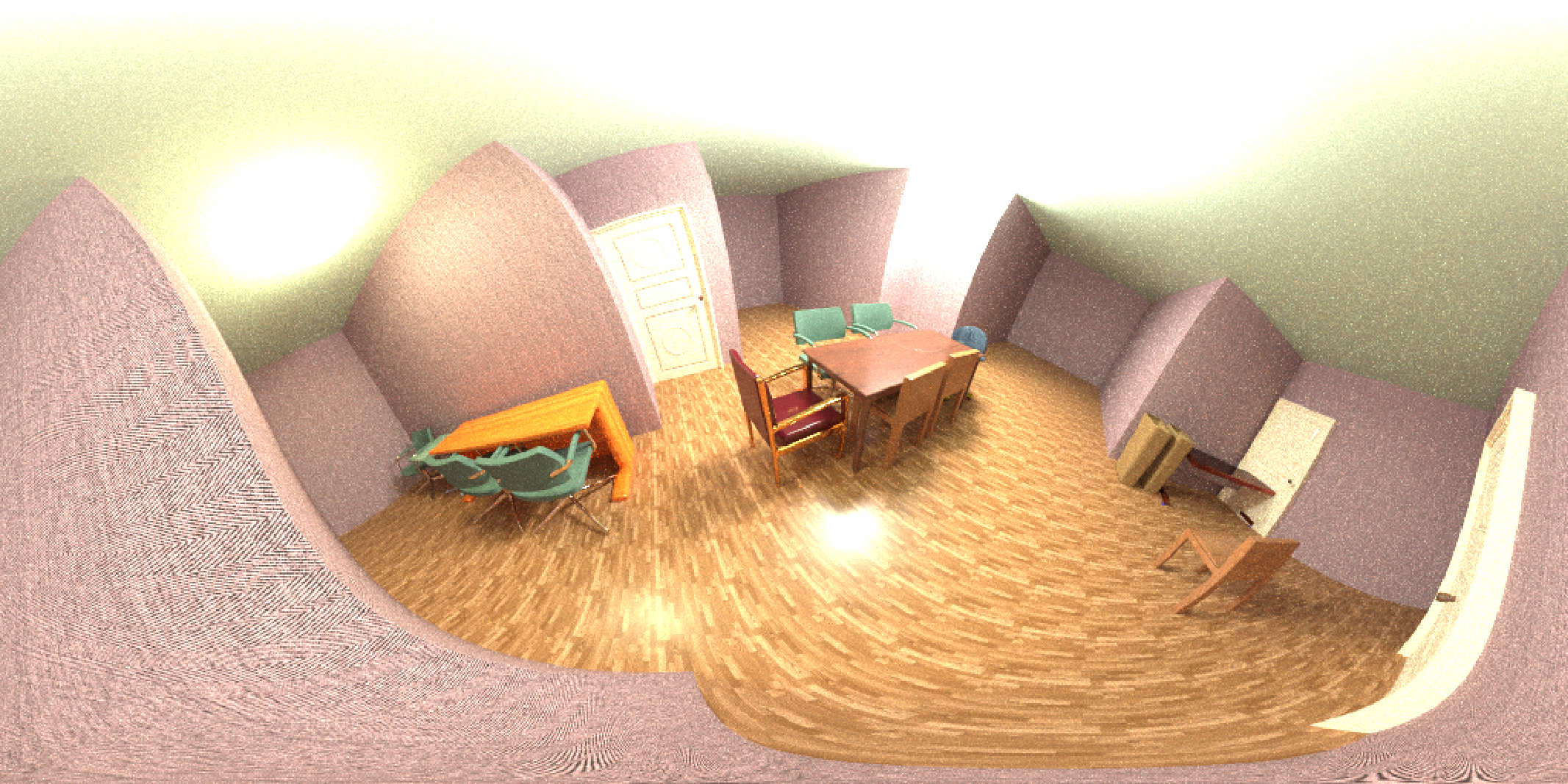} &
			\includegraphics[width=0.17\linewidth]{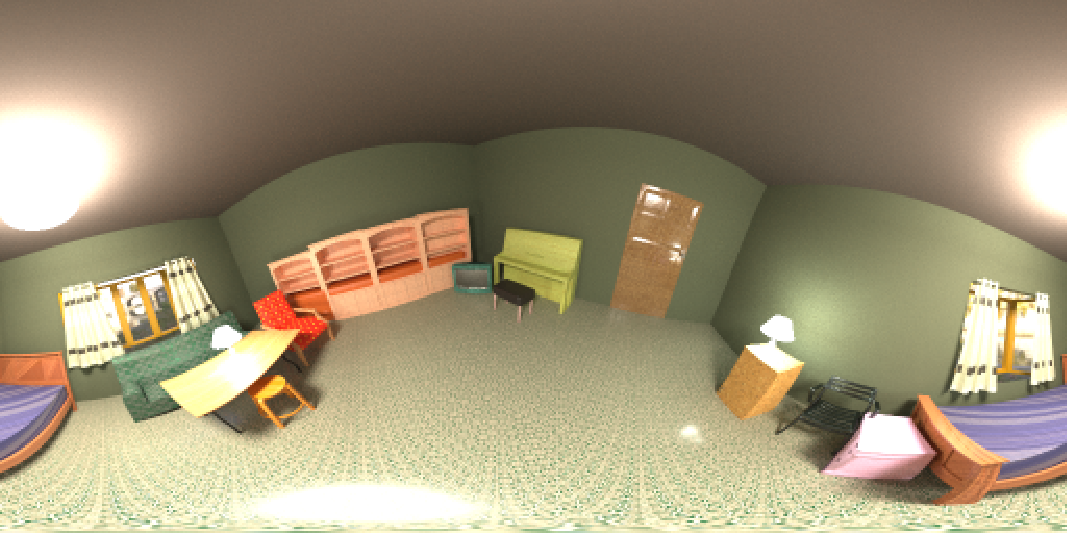} &
			\includegraphics[width=0.17\linewidth]{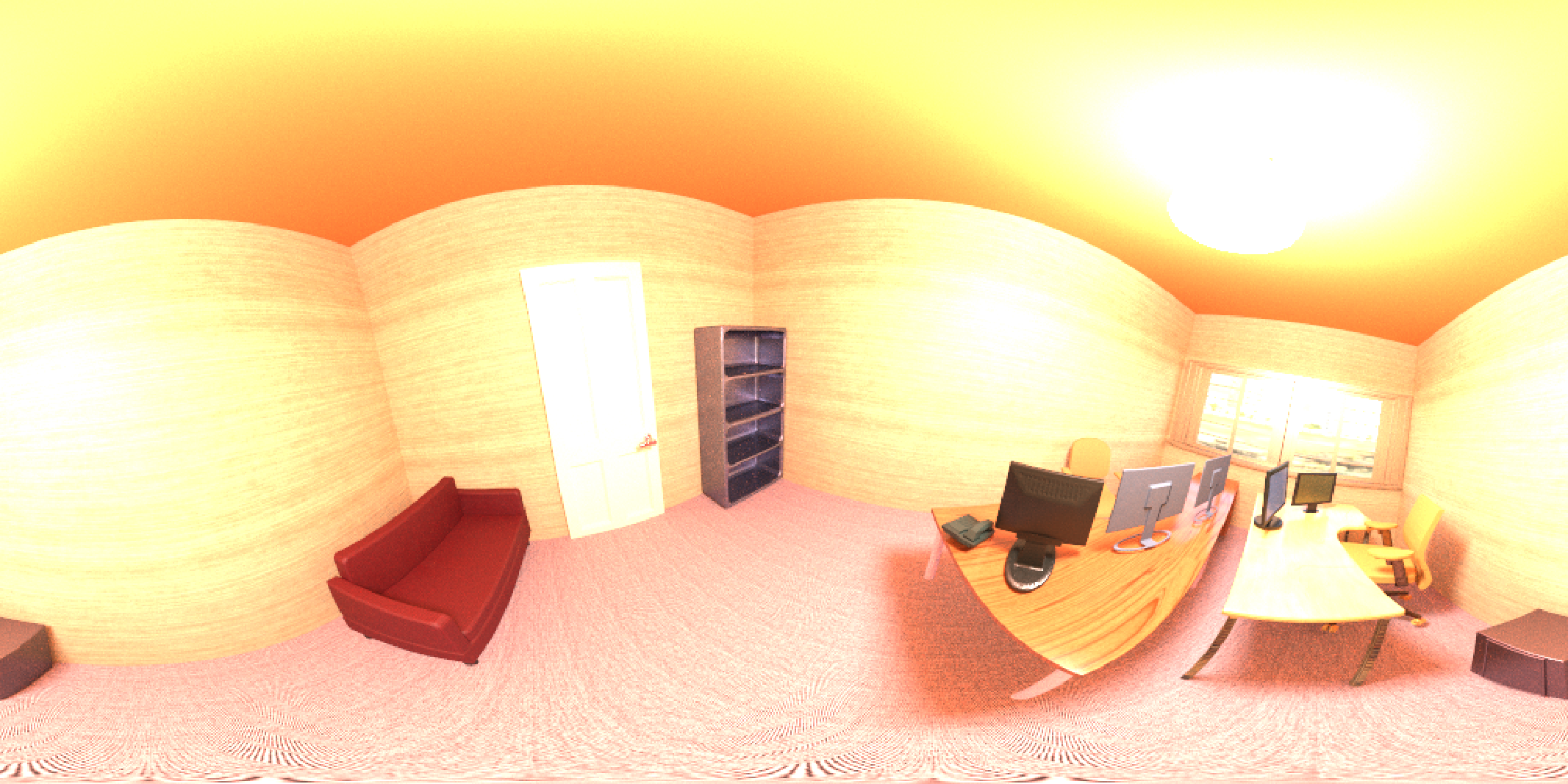} &
			\includegraphics[width=0.17\linewidth]{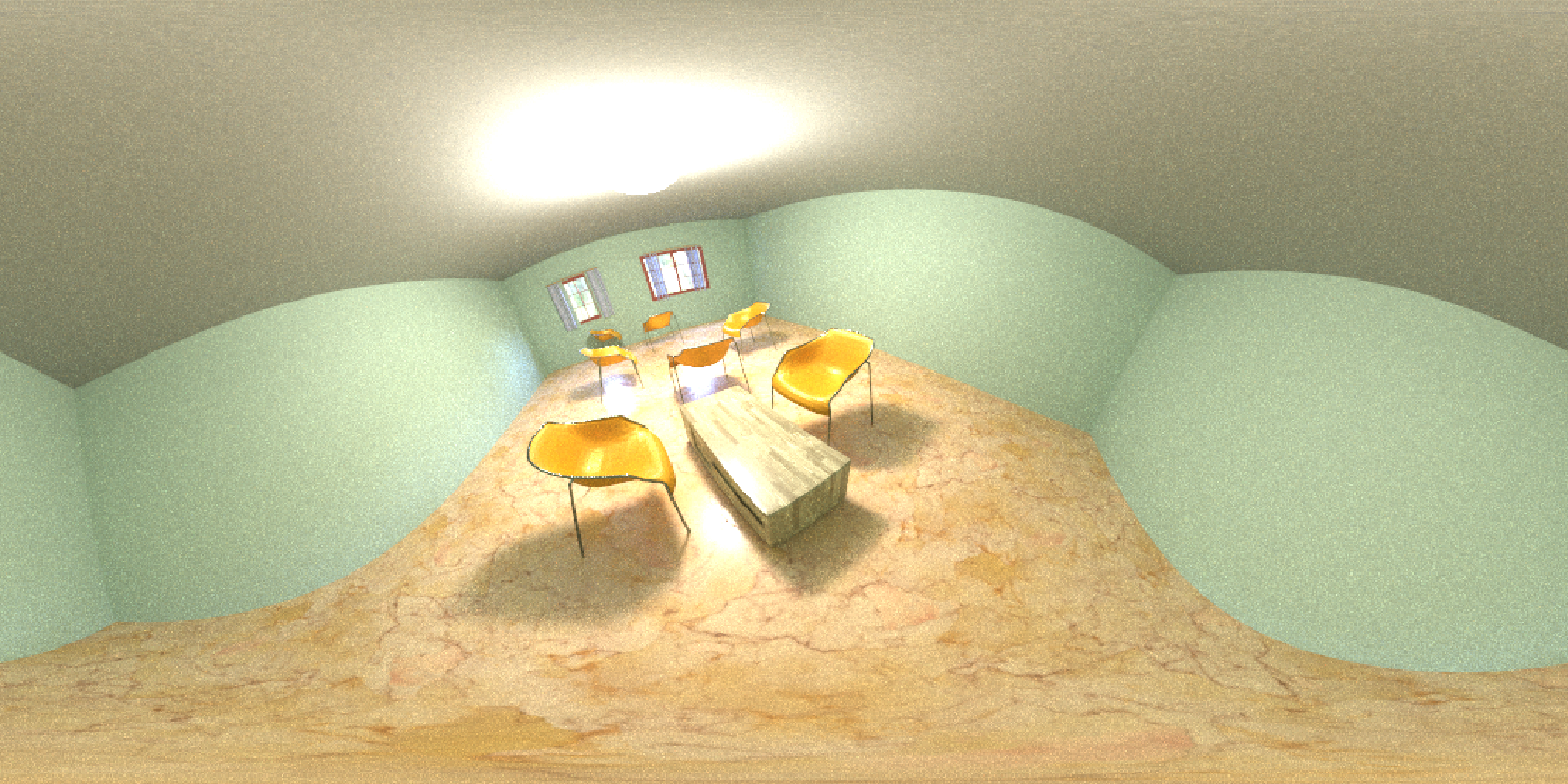}
			\\[0.1em]
		\end{tabular}
	\end{tabular}
	\caption{Illumination estimation results on our SpecularRooms dataset. We list the input reference frames of \textsf{Lighthouse} and the input point cloud of our method in the first row. %
		Our method retains more detailed information of a larger part of the scene.
	}
	\label{light_qualitative}
	\vspace{-0.9em}
\end{figure*}

\subsection{Our SpecularRooms Dataset}

Existing SLAM-related datasets, such as TUM-RGBD~\cite{sturm12tumrgbd} and EuRoC~\cite{burri2016euroc}, can not fulfill the requirements for evaluating non-Lambertian SLAM methods. They have at least one of the following limitations: 1) Incomplete ground truth for materials and illumination. 2) Lack of non-Lambertian surfaces. To address these limitations, we establish a new dataset that encompasses all the aforementioned criteria. We name this dataset ``SpecularRooms''. It contains more than 17000 images from 85 sequences of different lengths and complete ground truth trajectories, material, and illumination. The experiments of the following sections are conducted on our SpecularRooms dataset. In the experiments of material estimation and illumination estimation, we use 80\% of images for training and 20\% for testing. To validate our PBA method, we choose 6 longest trajectories. %

\subsection{Photometric Bundle Adjustment}

\noindent\textbf{Experimental Setup}.
Our pixel selection strategy adheres to the approach outlined in~\cite{engel2017direct}. The input trajectory is derived by estimating relative poses pairwise, which can be accomplished by any visual odometry or SLAM system, such as ORB-SLAM2~\cite{mur2017orb}.

\noindent\textbf{Methods for Comparison}.Here we refer to the classical PBA as \textsf{LBT-PBA} since it assumes a Lambertian world. We compare our \textsf{PB-PBA} with \textsf{LBT-PBA}~\cite{pba_cmu} and the state-of-the-art t-distribution based PBA work \textsf{DSM-PBA}~\cite{dsm}.
\begin{itemize}
	\item \textsf{LBT-PBA} is implemented based on the photometric consistency assumption, without considering non-Lambertian surfaces.
	\item \textsf{DSM-PBA} models the photometric error distribution through a selected t-distribution, thereby down-weighting outlier points.
\end{itemize}

\noindent\textbf{Metrics}. We follow~\cite{li2020robust_icra,wang2020unsupervised} to evaluate the trajectory accuracy based on the absolute trajectory error (ATE).

\noindent\textbf{Results}. As shown in Fig.~\ref{fig_trajectories}, \textsf{DSM-PBA} enhances the accuracy compared to the classical approach. Nonetheless, the precision of \textsf{DSM-PBA} is somewhat constrained due to its inability to adjust photometric error in a comprehensible manner. In contrast, our method attains superior accuracy, primarily because it incorporates a reflection model into the weight calculation process.
\begin{table}[!ht]
	\centering
	\renewcommand{\tabcolsep}{4.3pt} %
	\renewcommand{\arraystretch}{1.2}%
	\caption{Absolute Trajectory Errors of Various PBA Methods on
		Our SpecularRooms Dataset (Unit: meters)}
	\vspace{-0.5em}
	\label{tab:pba_metrics}
	\begin{tabular}{@{}lSS|SS|SS|SS|SS@{}}
		\toprule
		{~~Sequence }          & {\textsf{LBT-PBA}~\cite{pba_cmu}} & { \textsf{DSM-PBA}~\cite{dsm}} & {\textsf{PB-PBA}
		(our)}                                                                                                         \\
		\midrule
		\textit{Club}          & \textnormal{0.080}                & \textnormal{0.078}             & \textbf{0.074}   \\
		\textit{Office}        & \textnormal{0.093}                & \textnormal{0.088}             & \textbf{0.052}   \\
		\textit{Computer Room} & \textnormal{0.060}                & \textnormal{0.058}             & \textbf{0.032}   \\
		\textit{Meeting Room}  & \textnormal{0.105}                & \textnormal{0.069}             & \textbf{0.018}   \\
		\textit{Study Room}    & \textnormal{0.107}                & \textnormal{0.070}             & \textbf{0.056}   \\
		\textit{Restroom}      & \textnormal{0.115}                & \textnormal{0.062}             & \textbf{0.012}   \\
		\hline
		Average                & \textnormal{0.106}                & \textnormal{0.067}             & \textbf{0.031}   \\
		\bottomrule
	\end{tabular}
	\vspace{-0.9em}
\end{table}

\subsection{Material Estimation}

\noindent\textbf{Methods for Comparison}. We compare our method with \textsf{IRCIS}~\cite{li2020inverse}, a state-of-the-art work with single-frame input. To our knowledge, there is no open-source scene-level
method with multi-view input,
without requiring per-scene optimization. We re-train \textsf{IRCIS} on our SpecularRooms dataset using its recommended parameters for a fair comparison.

\noindent\textbf{Metrics}. Following~\cite{li2020inverse}, we use mean squared error (MSE) as our evaluation metric.

\begin{table}[!h]
	\centering
	\renewcommand{\tabcolsep}{4.3pt} %
	\renewcommand{\arraystretch}{1.15}%
	\caption{Quantatitive Results of Roughness Estimation on Our SpecularRooms Dataset}
	\label{tab:eval_metrics}
	\begin{tabular}{@{}lSS|SS}
		\toprule
		{~}                             & {\textsf{IRCIS}~\cite{li2020inverse}} & {\textsf{IRCIS+}} & {\textsf{Ours@5}} & {\textsf{Ours@10}} \\
		\midrule
		\text{Mean ($\times10^{-2}$)}   & \textnormal{2.00}                     & \textnormal{1.68} & \textnormal{1.46} & \textbf{1.26}      \\
		\text{Median ($\times10^{-2}$)} & \textnormal{0.72}                     & \textnormal{0.56} & \textnormal{0.55} & \textbf{0.50}      \\
		\bottomrule
	\end{tabular}
	\vspace{-0.9em}
\end{table}

\noindent\textbf{Results}. In Table~\ref{tab:eval_metrics}, we list the mean and median of per-frame MSE of our method with 5 frames (\textsf{Ours@5}) and 10 frames (\textsf{Ours@10}). \textsf{IRCIS+} denotes \textsf{IRCIS} with extra supervision information, e.g., depth, normal, albedo, etc. Some representative results are available in Fig.~\ref{fig_material}. Compared to \textsf{IRCIS}, our method with multi-view inputs achieves better results, since material's intrinsic properties are better demonstrated under varying viewpoints.

\subsection{Illumination Estimation}

\noindent\textbf{Methods for Comparison}. We compare our model with \textsf{Lighthouse}~\cite{lighthouse2020}. It uses a multiscale volumetric scene illumination representation, which can also produce spatial-consistent illumination results. For a fair comparison, we finetune the pre-trained model of \textsf{Lighthouse} on our SpecularRooms dataset with its recommended parameters.

\noindent\textbf{Metrics}. Following~\cite{lighthouse2020} and~\cite{legendre2019deeplight}, we report color angular error and peak signal-to-noise ratio (PSNR) for the predictions versus ground truth environment maps.

\noindent\textbf{Results}. Quantitative results are shown in Table~\ref{tab:light_metrics}. We also list some representative results in Fig.~\ref{light_qualitative}. Note that \textsf{Lighthouse} has a limited field of view, which means a large part of its prediction is from unseen areas and thus is unreliable. Our method performs better in most cases thanks to the larger visibility of the scene from the point cloud.

\begin{table}[!t]
	\centering
	\caption{Quantitative Results of Illumination Estimation on
		Our SpecularRooms Dataset}
	\vspace{-0.5em}
	\label{tab:light_metrics}
	\renewcommand{\arraystretch}{1.2}
	\begin{tabular}{@{}lSS|SS}
		\toprule
		                        & \multicolumn{2}{c|}{Angular Error ($^\circ$) $\downarrow$} & \multicolumn{2}{c}{PSNR~(dB) $\uparrow$}                                           \\
		                        & {\textsf{Lighthouse}~\cite{lighthouse2020}}                & {\textsf{Ours}}                          & {\textsf{Lighthouse}} & {\textsf{Ours}} \\
		\midrule
		\textit{Meeting Room}   & \textnormal{6.62}                                          & \textbf{2.44}                            & \textnormal{13.00}    & \textbf{19.93}  \\
		\textit{Computer Room}  & \textnormal{1.55}                                          & \textbf{1.09}                            & \textnormal{17.03}    & \textbf{19.22}  \\
		\textit{Office}         & \textnormal{8.58}                                          & \textbf{2.85}                            & \textnormal{13.54}    & \textbf{19.26}  \\
		\textit{Club}           & \textnormal{4.98}                                          & \textbf{1.67}                            & \textnormal{15.63}    & \textbf{21.84}  \\
		\textit{Apartment Room} & \textnormal{6.58}                                          & \textbf{2.30}                            & \textnormal{11.27}    & \textbf{19.22}  \\
		\hline
		Average                 & \textnormal{5.66}                                          & \textbf{2.07}                            & \textnormal{14.09}    & \textbf{19.89}  \\
		\bottomrule
	\end{tabular}
	\vspace{-0.9em}
\end{table}

\section{CONCLUSIONS}

In this paper, we introduce a novel PBA method for non-Lambertian environments. Our PBA method utilizes the material and illumination information to overcome the photometric inconsistency. The information is from our material and illumination estimation pipeline, which can effectively utilize multi-view input. Furthermore, we establish a new dataset to provide an evaluation framework for PBA methods in non-Lambertian environments.
Extensive experiments demonstrated that our PBA method
outperforms existing approaches in accuracy.

\bibliographystyle{IEEEbib_haoang}
\bibliography{ref_IEEEexample}

\begin{thebibliography}{10}

\bibitem{li2023hong}
H.~Li, J.~Zhao, J.-C. Bazin, P.~Kim, K.~Joo, Z.~Zhao, and Y.-H. Liu,
\newblock ``{Hong Kong World}: Leveraging structural regularity for line-based
  {SLAM},''
\newblock {\em TPAMI}, 2023.

\bibitem{wang2022efficient}
G.~Wang, X.~Wu, S.~Jiang, Z.~Liu, and H.~Wang,
\newblock ``Efficient {3D} deep {LiDAR} odometry,''
\newblock {\em TPAMI}, 2022.

\bibitem{huang2023learning}
T.~Huang, H.~Li, K.~He, C.~Sui, B.~Li, and Y.-H. Liu,
\newblock ``Learning accurate {3D} shape based on stereo polarimetric
  imaging,''
\newblock in {\em CVPR}, 2023.

\bibitem{li2020robust}
H.~Li, J.~Zhao, J.-C. Bazin, and Y.-H. Liu,
\newblock ``Robust estimation of absolute camera pose via intersection
  constraint and flow consensus,''
\newblock {\em TIP}, 2020.

\bibitem{engel2017direct}
J.~Engel, V.~Koltun, and D.~Cremers,
\newblock ``Direct sparse odometry,''
\newblock {\em TPAMI}, 2017.

\bibitem{zabih1994non}
R.~Zabih and J.~Woodfill,
\newblock ``Non-parametric local transforms for computing visual
  correspondence,''
\newblock in {\em ECCV}, 1994.

\bibitem{crivellaro2014robust}
A.~Crivellaro and V.~Lepetit,
\newblock ``Robust {3D} tracking with descriptor fields,''
\newblock in {\em CVPR}, 2014.

\bibitem{dai2017bundlefusion}
A.~Dai, M.~Nie{\ss}ner, M.~Zollh{\"o}fer, S.~Izadi, and C.~Theobalt,
\newblock ``Bundlefusion: Real-time globally consistent {3D} reconstruction
  using on-the-fly surface reintegration,''
\newblock {\em TOG}, 2017.

\bibitem{alismail2016direct}
H.~Alismail, M.~Kaess, B.~Browning, and S.~Lucey,
\newblock ``Direct visual odometry in low light using binary descriptors,''
\newblock {\em RAL}, 2016.

\bibitem{jin2003semi}
P.~Favaro, Hailin J., and S.~Soatto,
\newblock ``A semi-direct approach to structure from motion,''
\newblock in {\em ICIAP}, 2001.

\bibitem{klose2013efficient}
S.~Klose, P.~Heise, and A.~Knoll,
\newblock ``Efficient compositional approaches for real-time robust direct
  visual odometry from {RGB-D} data,''
\newblock in {\em IROS}, 2013.

\bibitem{engel2015large}
J.~Engel, J.~St{\"u}ckler, and D.~Cremers,
\newblock ``Large-scale direct {SLAM} with stereo cameras,''
\newblock in {\em IROS}, 2015.

\bibitem{scandaroli2012improving}
G.~Scandaroli, M.~Meilland, and R.~Richa,
\newblock ``Improving ncc-based direct visual tracking,''
\newblock in {\em ECCV}, 2012.

\bibitem{meilland2011real}
M.~Meilland, A.~Comport, and P.~Rives,
\newblock ``Real-time dense visual tracking under large lighting variations,''
\newblock in {\em BMVC}, 2011.

\bibitem{gonccalves2011real}
T.~Gon{\c{c}}alves and A.~Comport,
\newblock ``Real-time direct tracking of color images in the presence of
  illumination variation,''
\newblock in {\em ICRA}, 2011.

\bibitem{kerl2013robust}
C.~Kerl, J.~Sturm, and D.~Cremers,
\newblock ``Robust odometry estimation for {RGB-D} cameras,''
\newblock in {\em ICRA}, 2013.

\bibitem{dsm}
J.~Zubizarreta, I.~Aguinaga, and J.~Montiel,
\newblock ``Direct sparse mapping,''
\newblock {\em TRO}, 2020.

\bibitem{mur2017orb}
R.~Mur-Artal and J.~Tard{\'o}s,
\newblock ``{ORB-SLAM2}: An open-source {SLAM} system for monocular, stereo,
  and {RGB-D} cameras,''
\newblock {\em TRO}, 2017.

\bibitem{barron2013intrinsic}
J.~Barron and J.~Malik,
\newblock ``Intrinsic scene properties from a single {RGB-D} image,''
\newblock in {\em CVPR}, 2013.

\bibitem{barron2014shape}
J.~Barron and J.~Malik,
\newblock ``Shape, illumination, and reflectance from shading,''
\newblock {\em TPAMI}, 2015.

\bibitem{pbir_2023_ICCV}
C.~Sun, G.~Cai, Z.~Li, K.~Yan, C.~Zhang, C.~Marshall, J.~Huang, S.~Zhao, and
  Z.~Dong,
\newblock ``{Neural-PBIR} reconstruction of shape, material, and
  illumination,''
\newblock in {\em ICCV}, 2023.

\bibitem{zhu2022irisformer}
R.~Zhu, Z.~Li, J.~Matai, F.~Porikli, and M.~Chandraker,
\newblock ``Irisformer: Dense vision transformers for single-image inverse
  rendering in indoor scenes,''
\newblock in {\em CVPR}, 2022.

\bibitem{li2020inverse}
Z.~Li, M.~Shafiei, R.~Ramamoorthi, K.~Sunkavalli, and M.~Chandraker,
\newblock ``Inverse rendering for complex indoor scenes: Shape,
  spatially-varying lighting and {SVBRDF} from a single image,''
\newblock in {\em CVPR}, 2020.

\bibitem{jin_tensoir_2023}
H.~Jin, I.~Liu, P.~Xu, X.~Zhang, S.~Han, S.~Bi, X.~Zhou, Z.~Xu, and H.~Su,
\newblock ``{TensoIR}: Tensorial inverse rendering,''
\newblock in {\em CVPR}, 2023.

\bibitem{zhang_modeling_2022}
Y.~Zhang, J.~Sun, X.~He, H.~Fu, R.~Jia, and X.~Zhou,
\newblock ``Modeling indirect illumination for inverse rendering,''
\newblock in {\em CVPR}, 2022.

\bibitem{li2022texir}
Z.~Li, L.~Wang, M.~Cheng, C.~Pan, and J.~Yang,
\newblock ``Multi-view inverse rendering for large-scale real-world indoor
  scenes,''
\newblock in {\em CVPR}, 2023.

\bibitem{lighthouse2020}
P.~Srinivasan, B.~Mildenhall, M.~Tancik, J.~Barron, R.~Tucker, and N.~Snavely,
\newblock ``Lighthouse: Predicting lighting volumes for spatially-coherent
  illumination,''
\newblock in {\em CVPR}, 2020.

\bibitem{burri2016euroc}
M.~Burri, J.~Nikolic, P.~Gohl, T.~Schneider, J.~Rehder, S.~Omari, M.~W
  Achtelik, and R.~Siegwart,
\newblock ``The euroc micro aerial vehicle datasets,''
\newblock {\em IJRR}, 2016.

\bibitem{sturm12tumrgbd}
J.~Sturm, N.~Engelhard, F.~Endres, W.~Burgard, and D.~Cremers,
\newblock ``A benchmark for the evaluation of {RGB-D SLAM} systems,''
\newblock in {\em IROS}, Oct. 2012.

\bibitem{gardner2017learning}
M.~Gardner, K.~Sunkavalli, E.~Yumer, X.~Shen, E.~Gambaretto, C.~Gagn{\'e}, and
  J.~Lalonde,
\newblock ``Learning to predict indoor illumination from a single image,''
\newblock {\em arXiv}, 2017.

\bibitem{sengupta2019neural}
S.~Sengupta, J.~Gu, K.~Kim, G.~Liu, D.~Jacobs, and J.~Kautz,
\newblock ``Neural inverse rendering of an indoor scene from a single image,''
\newblock in {\em ICCV}, 2019.

\bibitem{legendre2019deeplight}
C.~LeGendre, W.~Ma, G.~Fyffe, J.~Flynn, L.~Charbonnel, J.~Busch, and
  P.~Debevec,
\newblock ``Deeplight: Learning illumination for unconstrained mobile mixed
  reality,''
\newblock in {\em CVPR}, 2019.

\bibitem{zhan2021sparse}
F.~Zhan, C.~Zhang, W.~Hu, S.~Lu, F.~Ma, X.~Xie, and L.~Shao,
\newblock ``Sparse needlets for lighting estimation with spherical transport
  loss,''
\newblock in {\em ICCV}, 2021.

\bibitem{li2023spatiotemporally}
Z.~Li, L.~Yu, M.~Okunev, M.~Chandraker, and Z.~Dong,
\newblock ``Spatiotemporally consistent {HDR} indoor lighting estimation,''
\newblock {\em TOG}, 2023.

\bibitem{Cook1982}
R.~Cook and K.~Torrance,
\newblock ``A reflectance model for computer graphics,''
\newblock {\em TOG}, 1982.

\bibitem{levin1998approximation}
D.~Levin,
\newblock ``The approximation power of moving least-squares,''
\newblock {\em Mathematics of computation}, 1998.

\bibitem{li2018monocular}
H.~Li, J.~Yao, J.-C. Bazin, X.~Lu, Y.~Xing, and K.~Liu,
\newblock ``A monocular {SLAM} system leveraging structural regularity in
  {Manhattan} world,''
\newblock in {\em ICRA}, 2018.

\bibitem{achanta2012slic}
R.~Achanta, A.~Shaji, K.~Smith, A.~Lucchi, P.~Fua, and S.~S{\"u}sstrunk,
\newblock ``{SLIC} superpixels compared to state-of-the-art superpixel
  methods,''
\newblock {\em TPAMI}, 2012.

\bibitem{li2019leveraging}
H.~Li, Y.~Xing, J.~Zhao, J.-C. Bazin, Z.~Liu, and Y.-H. Liu,
\newblock ``Leveraging structural regularity of {Atlanta} world for monocular
  {SLAM},''
\newblock in {\em ICRA}, 2019.

\bibitem{wang2021end}
Y.~Wang, Z.~Xu, X.~Wang, C.~Shen, B.~Cheng, H.~Shen, and H.~Xia,
\newblock ``End-to-end video instance segmentation with transformers,''
\newblock in {\em CVPR}, 2021.

\bibitem{wang2004ssim}
Z.~Wang, A.~Bovik, H.~Sheikh, and E.~Simoncelli,
\newblock ``Image quality assessment: from error visibility to structural
  similarity,''
\newblock {\em TIP}, 2004.

\bibitem{pba_cmu}
H.~Alismail, B.~Browning, and S.~Lucey,
\newblock ``Photometric bundle adjustment for vision-based {SLAM},''
\newblock in {\em ACCV}, 2017.

\bibitem{qi2017pointnet++}
C.~Qi, L.~Yi, H.~Su, and L.~Guibas,
\newblock ``Pointnet++: Deep hierarchical feature learning on point sets in a
  metric space,''
\newblock {\em NeurIPS}, 2017.

\bibitem{synpt2020}
Z.~Song, W.~Chen, D.~Campbell, and H.~Li,
\newblock ``Deep novel view synthesis from colored {3D} point clouds,''
\newblock in {\em ECCV}, 2020.

\bibitem{li2020robust_icra}
H.~Li, W.~Chen, J.~Zhao, J.-C. Bazin, L.~Luo, Z.~Liu, and Y.-H. Liu,
\newblock ``Robust and efficient estimation of absolute camera pose for
  monocular visual odometry,''
\newblock in {\em ICRA}, 2020.

\bibitem{wang2020unsupervised}
G.~Wang, C.~Zhang, H.~Wang, J.~Wang, Y.~Wang, and X.~Wang,
\newblock ``Unsupervised learning of depth, optical flow and pose with
  occlusion from {3D} geometry,''
\newblock {\em TITS}, 2020.

\end{thebibliography}

\end{document}